\documentclass[pdflatex,sn-mathphys-num]{sn-jnl}

\usepackage{graphicx}%
\usepackage{multirow}%
\usepackage{amsmath,amssymb,amsfonts}%
\usepackage{amsthm}%
\usepackage{mathrsfs}%
\usepackage[title]{appendix}%
\usepackage{textcomp}%
\usepackage{manyfoot}%
\usepackage{booktabs}%
\usepackage{algorithm}%
\usepackage{algorithmicx}%
\usepackage{algpseudocode}%
\usepackage{listings}%

\usepackage{array}
\usepackage{makecell}
\usepackage{adjustbox}
\usepackage[table]{xcolor}
\usepackage{longtable}
\usepackage{subcaption}

\definecolor{FedAvgColor}{RGB}{228,74,69}
\definecolor{FedSelectColor}{RGB}{139,195,91}
\definecolor{IOPFLColor}{RGB}{126,94,177}
\definecolor{FedCorrColor}{RGB}{245,180,64}

\newcolumntype{L}[1]{>{\raggedright\arraybackslash}p{#1}}
\newcolumntype{C}[1]{>{\centering\arraybackslash}p{#1}}
\newcolumntype{M}[1]{>{\centering\arraybackslash}m{#1}}

\newcommand{\datasetthumb}[1]{%
  \adjustbox{valign=t}{%
    \includegraphics[width=2.2cm,height=4.4cm,keepaspectratio]{#1}%
  }%
}

\theoremstyle{thmstyleone}%
\theoremstyle{thmstyletwo}%

\theoremstyle{thmstylethree}%

\begin{document}

\title[Article Title]{Federated Medical Image Segmentation under Real-World Label Noise: A Benchmark Suite for Noisy Label Learning Method Selection}


\author*[1,2]{\fnm{Markus} \sur{Bujotzek}}\email{markus.bujotzek@dkfz-heidelberg.de}
\author[1]{\fnm{Dimitrios} \sur{Bounias}}
\author[1]{\fnm{Stefan} \sur{Denner}}
\author[1,3]{\fnm{Ralf} \sur{Floca}}
\author[1,2]{\fnm{Maximilian} \sur{Fischer}}
\author[1,4]{\fnm{Peter} \sur{Neher}}
\author[1,2,4,5]{\fnm{Klaus} \sur{Maier-Hein}}

\affil[1]{\orgdiv{Division of Medical Image Computing}, \orgname{Germany Cancer Research Center}, \city{Heidelberg}, \postcode{69120}, \country{Germany}}

\affil[2]{\orgdiv{Medical Faculty}, \orgname{University of Heidelberg}, \city{Heidelberg}, \postcode{69120}, \country{Germany}}

\affil[3]{\orgdiv{Heidelberg Institute of Radiation Oncology (HIRO)}, \orgname{National Center for Radiation Research in Oncology (NCRO)}, \city{Heidelberg}, \postcode{69120}, \country{Germany}}

\affil[4]{\orgdiv{Pattern Analysis and Learning Group, Department of Radiation Oncology}, \orgname{Heidelberg University Hospital}, \city{Heidelberg}, \postcode{69120}, \country{Germany}}

\affil[5]{\orgdiv{Faculty of Mathematics and Computer Science}, \orgname{University of Heidelberg}, \city{Heidelberg}, \postcode{69120}, \country{Germany}}

\affil[6]{\orgdiv{National Center for Tumor Diseases (NCT)}, \orgname{NCT Heidelberg, a partnership between DKFZ and the university medical center Heidelberg}, \city{Heidelberg}, \postcode{69120}, \country{Germany}}




\abstract{
\textbf{Objective:} 
While federated learning (FL) enables collaborative medical image segmentation without centralizing sensitive data, real-world deployment is frequently complicated by cross-site label imperfections such as contour disagreement, missing or additional structures, and confused labels.
Federated noisy label learning (FNLL) aims to mitigate these effects, yet remains underused in practice as existing evidence is largely based on synthetic noise, simplified settings, and limited real-world noisy evaluation.
We address this gap by introducing a benchmark suite that combines diverse real-world noisy datasets, deployment-relevant client-noise scenarios, and label-noise-targeted evaluation to support systematic FNLL assessment and informed method selection.

\textbf{Materials \& Methods:} 
The suite combines curated real-world noisy medical image segmentation datasets from diverse sources with a comprehensive federated segmentation framework including various client-noise scenarios and noise-targeted evaluation.
To demonstrate its capabilities, we compare representative FNLL methods across approaches, including noise-aware aggregation, robust personalization, label correction, and sample selection.

\textbf{Results:} 
In-depth data analysis shows that real-world segmentation label noise occurs both in isolation and in combination of characterized noise types.
The benchmark identifies \textit{FedSelect} as the strongest overall FNLL method, underlines \textit{FedAvg} as a competitive baseline, and provides an actionable decision guide to support selection of suitable FNLL strategies based on label-noise type and client-noise scenario.

\textbf{Discussion \& Conclusion:} 
The presented suite provides a realistic and discriminative basis for FNLL evaluation in medical image segmentation and establishes a reusable foundation for fair benchmarking, dataset-specific label-noise characterization, and future method development under realistic federated settings. Code is available at \hyperlink{https://github.com/MIC-DKFZ/FedSegNoiseBench}{https://github.com/MIC-DKFZ/FedSegNoiseBench}.}

\keywords{Federated Learning; Medical Image Segmentation; Label Noise; Data Quality; Benchmarking}



\maketitle
\section{Introduction}
\label{sec:intro}
Federated Learning (FL) enables collaborative model training across institutions without exchanging local data, by aggregating locally optimized model updates \cite{rieke2020future}. 
In medical imaging, FL leverages multi-institutional data quantity and diversity to improve robustness and generalization while respecting privacy and governance constraints \cite{Sheller2020FederatedLI}, with growing translational evidence in radiology \cite{pati2022federated, bujotzek2025real} and beyond \cite{oldenhof2023industry}.

\subsection{Motivation}
\begin{figure}[htp]
\centering
\includegraphics[width=\textwidth]{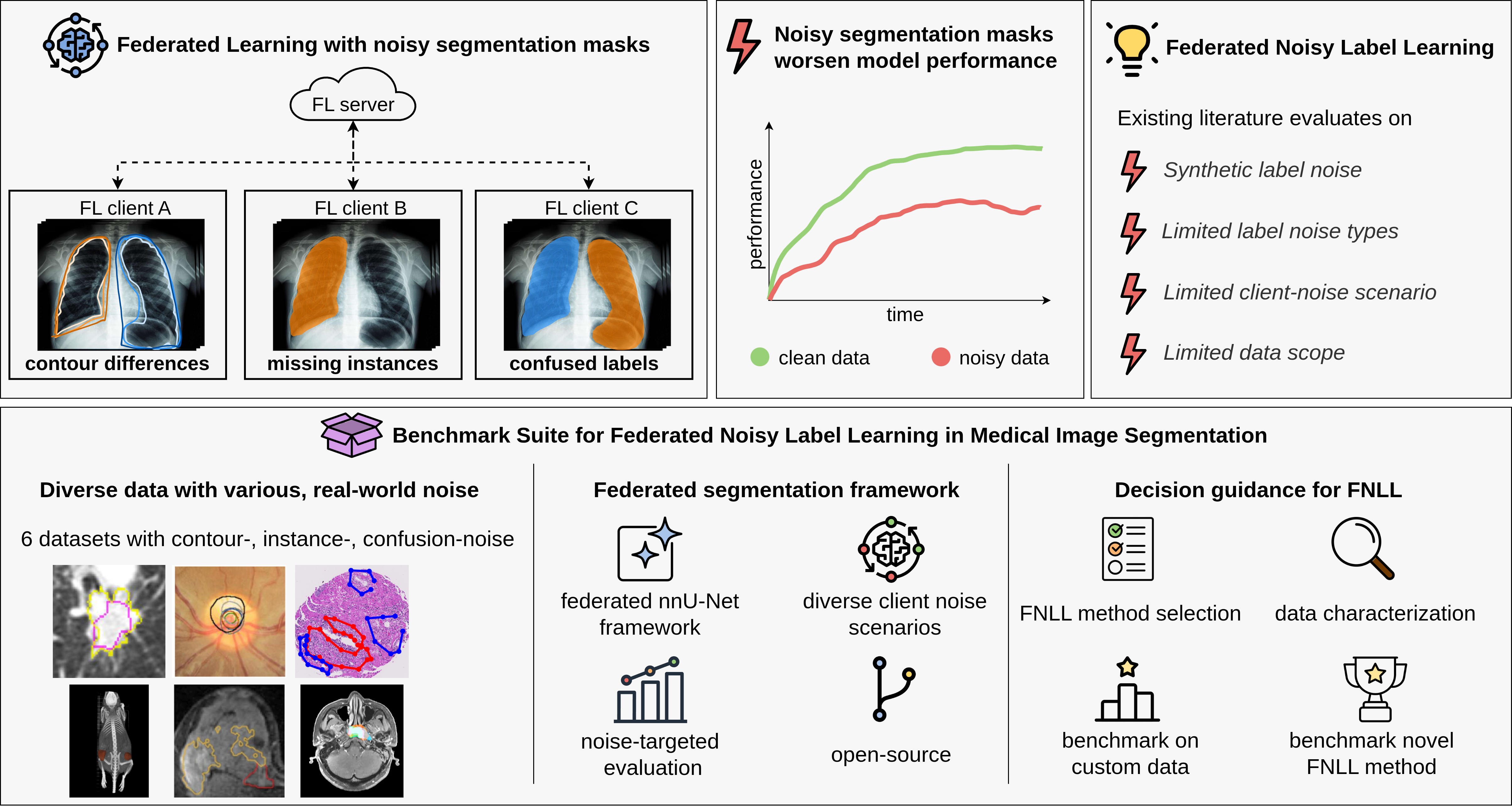}
\caption{Segmentation label noise of various forms degrades model performance and poses a particular challenge in FL, where noisy annotations are distributed across clients and cannot be centrally inspected. While FNLL methods aim to address this problem, existing literature is often limited to few and synthetic noise types, restricted client-noise scenarios, and narrow data scope. Our benchmark suite closes this gap by combining diverse real-world noisy segmentation datasets, a federated benchmarking framework, and comprehensive noise-targeted evaluation, thereby enabling FNLL method selection, dataset characterization, benchmarking on new data, and evaluation of newly developed FNLL methods. Dataset thumbnails adapted from \cite{armato2011lung, liao2024modeling, zhong2017curated, jensen20243d, wu2024dataset, garrucho2025}.}
\label{fig:fig1}
\end{figure}

Medical image segmentation is a key component of automated diagnosis and computer-assisted therapy, as it provides spatially meaningful representations of anatomical and pathological structures \cite{antonelli2022medical}.
Supervised training of segmentation models relies on voxel-level annotation masks, which in clinical practice are often heterogeneous and noisy due to inter-rater variability, difficulty of target recognition, human annotation errors, and the increasing use of automatically generated labels \cite{karimi2020deep}.
As a result, segmentation label noise commonly manifests as contour inconsistencies, missing or additional instances, or confused class labels of targets, all of which degrade model performance \cite{karimi2020deep} (Figure~\ref{fig:fig1}).
Because segmentation outputs are frequently used in downstream tasks such as diagnosis, treatment planning, and quantitative biomarker extraction, such degradation can propagate into clinical workflows and affect decision-making reliability \cite{aliotta2019quantifying, poel2022impact, ma2024segment}.
These noise-based challenges are further amplified in FL.
First, both data and annotation heterogeneity increase naturally as image acquisition, curation, and labeling are performed independently across institutions \cite{fang2022robust}.
Second, the privacy-preserving nature of FL restricts centralized inspection of data and labels, complicating noise detection and mitigation, such that noisy clients may only become apparent through performance deterioration, as reported in real-world deployments \cite{bujotzek2025real}.
Third, distributed label noise can propagate through model aggregation, allowing partially or fully noisy clients to degrade learning on otherwise clean clients.

\subsection{Related work and problem statement}
Centralized Noisy Label Learning (NLL) is a widely studied domain beyond medical image segmentation \cite{karimi2020deep, shi2024survey, wei2024deep, song2022learning}, and methods can be categorized into model- and data-centric approaches.
Model-centric approaches include noise-robust architecture \cite{dgani2018training} or loss \cite{hu2021deep, chen2022adaptive} modifications, regularization via label smoothing \cite{pham2021interpreting} or data augmentation \cite{qian2025adaptive}, and changed training paradigms like curriculum learning \cite{zhou2020robust} or teacher-student frameworks \cite{qian2025adaptive, li2023semi}.
Data-centric approaches cover approaches like label correction or refinement \cite{zheng2021meta, huang2022contrastive, zhou2020robust, qian2025adaptive}, data re-weighting \cite{ma2021learning, li2019learning, mirikharaji2019learning, xiao2021pathological}, or sample selection \cite{mindermann2022prioritized, xia2021sample, kim2021fine, ribeiro2020less, qian2025adaptive, li2023semi}.
While centralized NLL has been extensively studied, many methods rely on pooled-data signals, like global loss statistics, class centroids, or clean validation sets, which are unavailable in cross-silo FL.
Moreover, client-specific data and annotation heterogeneity couple label reliability to domain shift \cite{xu2022fedcorr, zhang2025proxy}, making local noise mitigation prone to miscalibration across clients and potentially biasing global aggregation.

Federated Noisy Label Learning (FNLL) mitigates label noise under distributed-data constraints through four recurring, often combined mechanisms:
\textit{Noise-aware federated model aggregation} estimates client reliability and reweights updates during aggregation, typically using training dynamics, class or region losses, or class centroids to down-weight noisy clients \cite{wu2023fednoro, jiang2024tackling, chen2023medical}. 
In segmentation, aggregation can additionally be guided by contour-localized losses or completeness proxies \cite{wu2024feda3i, xiang2024fedia, zhu2023feddm, wang2025fedhnr}.
\textit{Noise-robust personalization} addresses data and label heterogeneity by adapting parts of the model to client-specific distributions. Methods use inter-site prediction inconsistency to retain local parameters and reweight uncertain regions \cite{wang2023feddp}, or fuse accumulated local-gradient information with the global model after training \cite{jiang2023iop}.
\textit{Label correction} methods identify unreliable labels, typically on noisy clients, and refine them using global predictions \cite{xu2022fedcorr}, centroid-consistent pseudo-labels \cite{yang2022robust}, or soft-label optimization \cite{jiang2024tackling}. 
In segmentation, correction is commonly performed pixel-wise through peer- or teacher-model agreement \cite{zhu2023feddm, bai2024improving}, or by confidence-thresholded recovery of missing structures \cite{xiang2024fedia}.
\textit{Sample selection} methods prioritize informative or reliable samples during training, either by focusing on high-loss (not-yet-learned) samples \cite{shin2022fedbalancer} or by mitigating label noise through confidence- and consistency-based selection \cite{yang2022robust}, including meta-learned strategies driven by loss dynamics \cite{zhang2025proxy}.

While FNLL methods have been proposed across diverse approaches, their evaluation remains fragmented along critical axes. 
First, studies predominantly rely on \textit{few} and largely \textit{synthetic noise} models \cite{wu2023fednoro, wu2024feda3i, xiang2024fedia, zhu2023feddm, jiang2024tackling, chen2023medical, bai2024improving, xu2022fedcorr, yang2022robust, zhang2025proxy}, despite evidence that conclusions may not transfer to real-world noisy data \cite{ma2025benchmarking}.  
Second, only \textit{limited client-noise scenarios} are considered \cite{wu2023fednoro, jiang2024tackling, chen2023medical, wu2024feda3i, xiang2024fedia, zhu2023feddm, wang2025fedhnr, xu2022fedcorr, bai2024improving}, rather than systematically assessing robustness under heterogeneous federated conditions.
Third, evaluations are typically conducted on a \textit{small number of narrow datasets} with limited modality diversity \cite{wu2023fednoro, jiang2024tackling, chen2023medical, wu2024feda3i, xiang2024fedia, zhu2023feddm, wang2025fedhnr, wang2023feddp, jiang2023iop, xu2022fedcorr, yang2022robust, bai2024improving, shin2022fedbalancer, zhang2025proxy}, restricting generalization.
As a result, FNLL remains underutilized in practice, where real-world FL deployments still rely on FedAvg-based training \cite{pati2022federated, bujotzek2025real, moradi2025beyond}, while evidence for FNLL is largely confined to synthetic and simplified settings.

Benchmarking endeavors have helped consolidate otherwise fragmented NLL research in centralized medical image classification \cite{ma2025benchmarking} and instance segmentation \cite{dalva2023benchmarking}, and have recently emerged for FNLL in classification \cite{jiang2025fnbench, liang2023fednoisy}.
However, a benchmarking effort for federated medical image segmentation systematically covering \textit{diverse datasets}, \textit{multiple client-noise scenarios}, and \textit{numerous, real-world} segmentation label noise types remains lacking, limiting informed FNLL method evaluation and selection.

\subsection{Contribution}
We present a reusable and extensible benchmark suite for FNLL in cross-silo medical image segmentation, enabling systematic evaluation, fair comparison, and informed method selection under realistic conditions.
Rather than proposing another narrowly evaluated FNLL method, we provide a reusable community resource for deployment-relevant evaluation, decision guidance, and structured extension to future datasets and methods.
The suite combines four tightly integrated components: 
(1) six curated medical image segmentation datasets with inherent real-world label noise and corresponding noise-type analysis;
(2) an nnU-Net-based federated segmentation framework \cite{isensee2021nnu, isensee2024nnu} with integrated state-of-the-art FNLL methods and deployment-relevant client-noise scenarios;
(3) a label-noise-targeted evaluation protocol for comparison across heterogeneous datasets and noise settings; and
(4) an actionable decision guide linking observed label-noise characteristics to suitable FNLL strategies.

Using this suite, we benchmark representative FNLL methods to identify consistently strong performers, assess robustness across datasets and noise scenarios, and characterize method suitability for specific label-noise patterns. By releasing the suite as an open-source codebase along a detailed contribution guide, we provide a community resource for reproducible evaluation, dataset-specific noise characterization, and structured extension to future datasets and methods. The code is publicly available at \hyperlink{https://github.com/MIC-DKFZ/FedSegNoiseBench}{https://github.com/MIC-DKFZ/FedSegNoiseBench}.

\section{Materials and Methods}\label{sec:materialsnmethods}

\subsection{Medical image segmentation datasets with real-world label noise}
\label{sec2_data}
We base our benchmark suite on six multi-rater medical image segmentation datasets with inherent inter-rater variability and real-world label noise, rather than synthetically generated noise.
Noisy training labels are obtained by randomly selecting one rater annotation per sample \cite{chen2023medical}, while clean labels are derived from majority voting, STAPLE fusion \cite{warfield2004simultaneous}, or additionally provided expert labels.

The curated datasets cover diverse clinical domains and modalities, including thoracic CT (LIDC-IDRI \cite{armato2011lung}), retinal fundus photography (RIGA \cite{almazroa2018retinal}), prostate histopathology microscopy (GleasonXAI Harvard Dataverse \cite{zhong2017curated, arvaniti2018automated, mittmann2025pathologist}), micro-CT of mouse tumors (MouseTumor \cite{jensen20243d}), and multi-organ MR imaging (MMIS \cite{wu2024dataset}, MAMA-MIA \cite{garrucho2025}).
They vary in dimensionality, number of classes and instances per class, and label-noise types, including contour inconsistencies, missing or additional target instances, and confused class labels, providing a broad test bed for FNLL in medical image segmentation.
Dataset and label-noise details are summarized in Table~\ref{tab:datasets}.

\begin{table*}[t]
\centering

\caption{Overview of the six benchmark datasets, their inherent real-world segmentation label-noise characteristics, and representative thumbnails qualitatively illustrating the source and manifestation of label noise. Example thumbnails adapted from \cite{armato2011lung, liao2024modeling, zhong2017curated, jensen20243d, wu2024dataset, garrucho2025}.}

\footnotesize
\setlength{\tabcolsep}{4pt}
\renewcommand{\arraystretch}{1.12}

\begin{adjustbox}{max width=\textwidth}
\begin{tabular}{
    L{2.0cm}
    *{6}{L{1.98cm}}
}
\toprule
& 
\multicolumn{1}{c}{\textbf{LIDC}} &
\multicolumn{1}{c}{\textbf{RIGA}} &
\multicolumn{1}{c}{\textbf{GleasonHD}} &
\multicolumn{1}{c}{\textbf{MouseT}} &
\multicolumn{1}{c}{\textbf{MMIS}} &
\multicolumn{1}{c}{\textbf{MMIA}} \\
\midrule

\textit{Modality}
& \centering CT
& \centering Fundus
& \centering Microscopy
& \centering micro-CT
& \centering MR
& \centering MR \tabularnewline

\textit{Dimensions}
& \centering 3D
& \centering 2D
& \centering 2D
& \centering 3D
& \centering 3D
& \centering 3D \tabularnewline

\textit{\# classes}
& \centering 1
& \centering 2
& \centering 3
& \centering 1
& \centering 1
& \centering 1 \tabularnewline

\midrule

\textit{Noise origin}
& \makecell[tl]{multi-rater}
& \makecell[tl]{multi-rater}
& \makecell[tl]{multi-rater}
& \makecell[tl]{multi-rater}
& \makecell[tl]{multi-rater}
& \makecell[tl]{auto-generated} \tabularnewline

\textit{Noise type}
& \makecell[tl]{contour,\\missed/extra\\labels}
& contour
& \makecell[tl]{contour,\\missed/extra\\labels,\\confused\\labels}
& \makecell[tl]{contour,\\missed/extra\\labels}
& \makecell[tl]{contour,\\missed/extra\\labels}
& \makecell[tl]{contour,\\missed/extra\\labels} \tabularnewline

\textit{Clean origin}
& \makecell[tl]{rater\\majority}
& \makecell[tl]{rater\\majority}
& STAPLE
& STAPLE
& \makecell[tl]{rater\\majority}
& expert \tabularnewline

\midrule

\textit{Example}
& \multicolumn{1}{c}{\datasetthumb{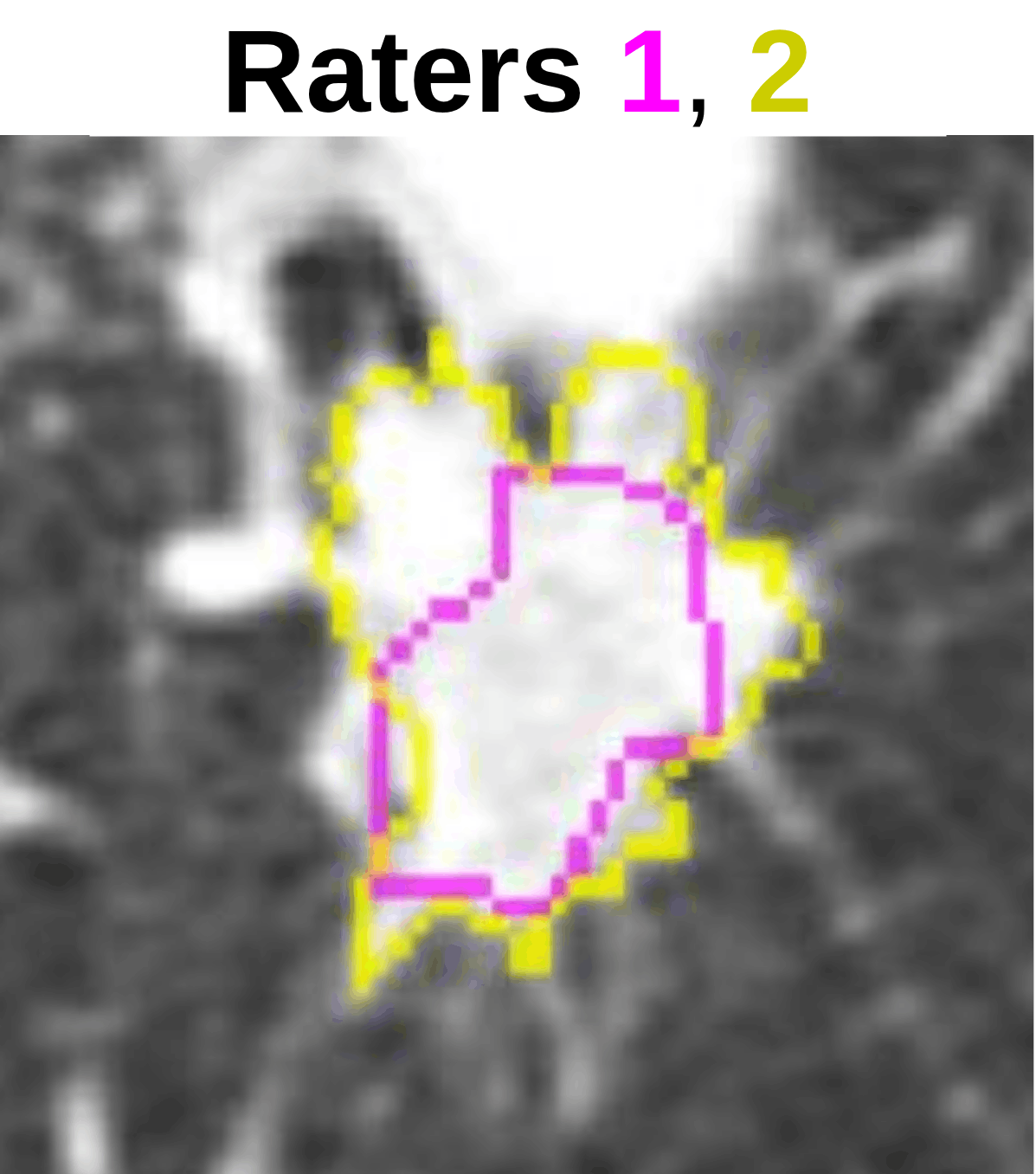}}
& \multicolumn{1}{c}{\datasetthumb{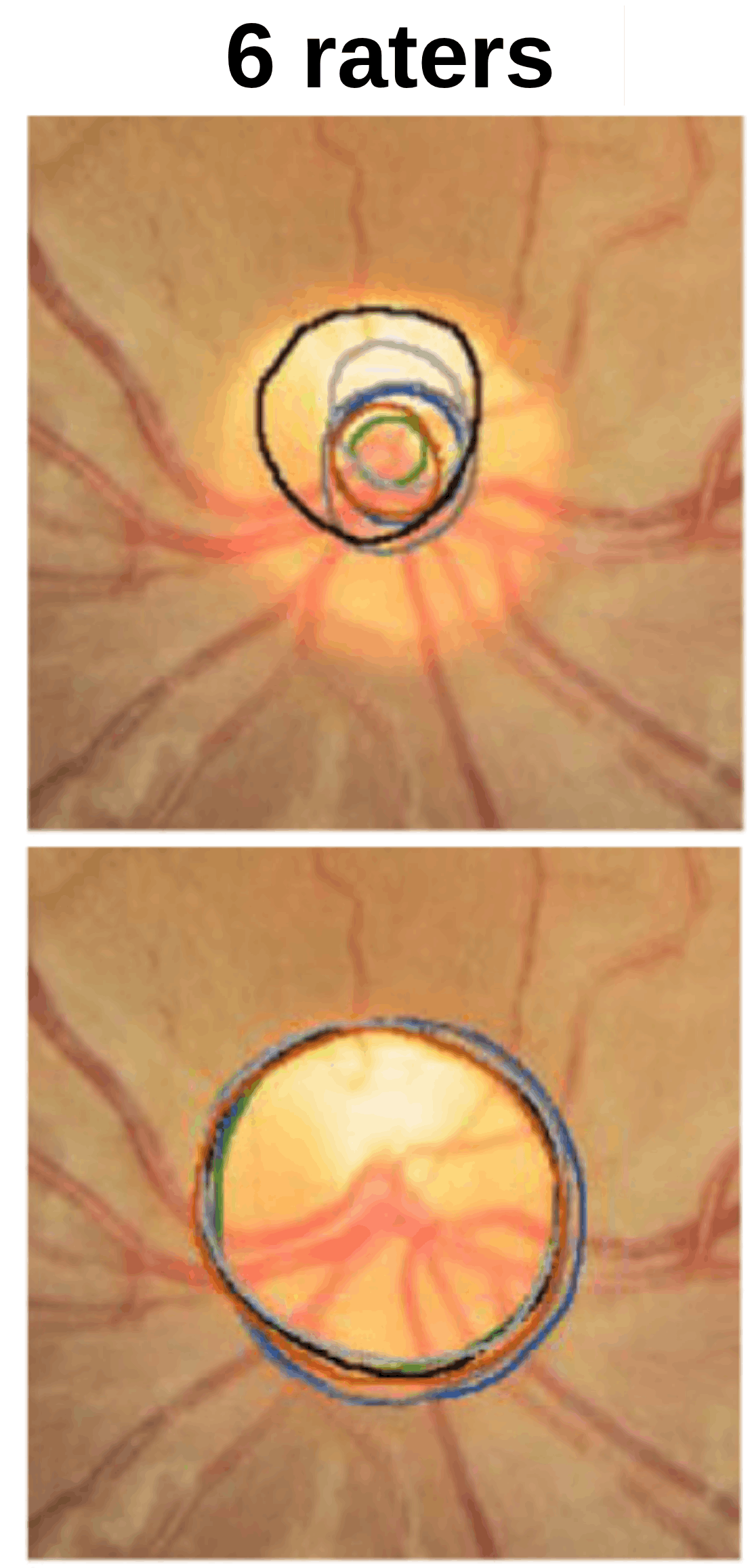}}
& \multicolumn{1}{c}{\datasetthumb{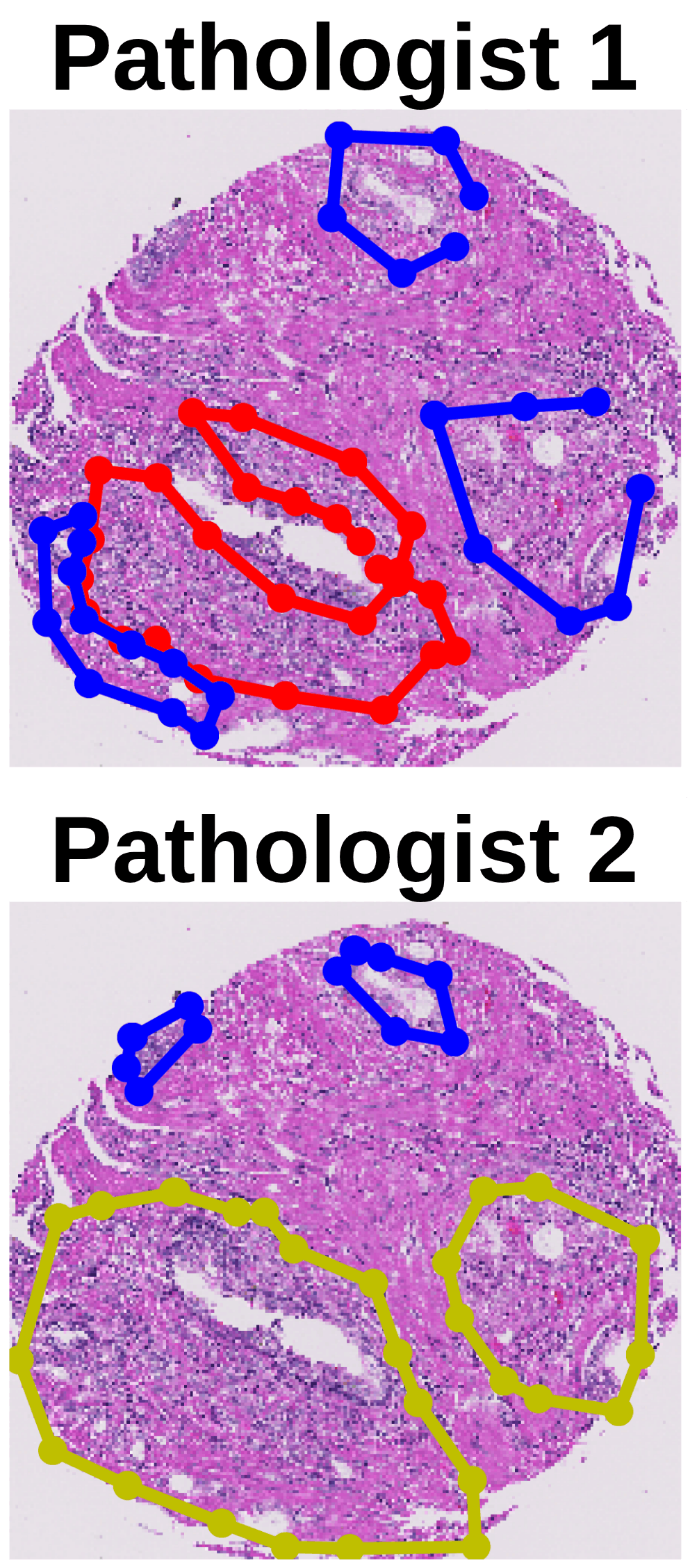}}
& \multicolumn{1}{c}{\datasetthumb{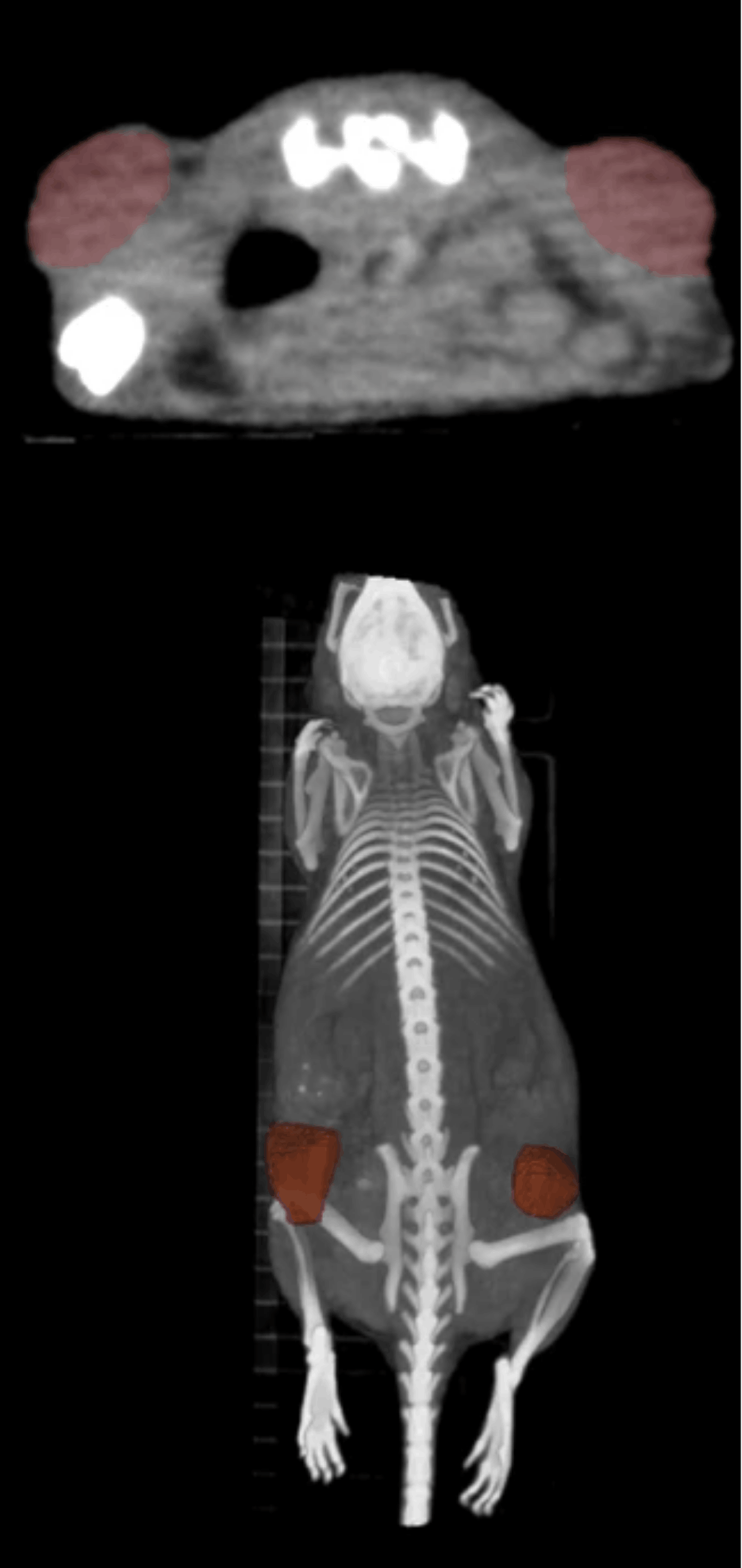}}
& \multicolumn{1}{c}{\datasetthumb{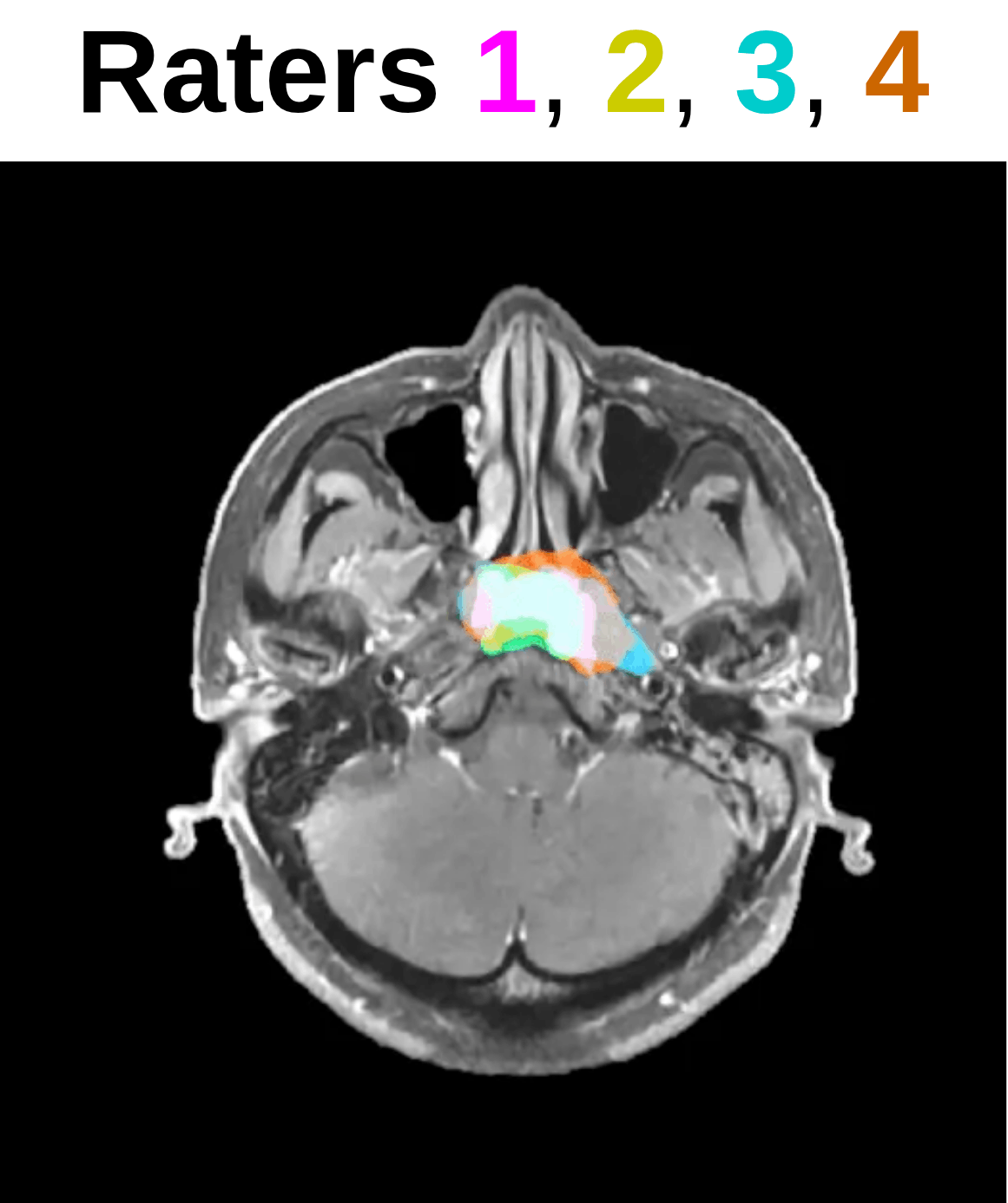}}
& \multicolumn{1}{c}{\datasetthumb{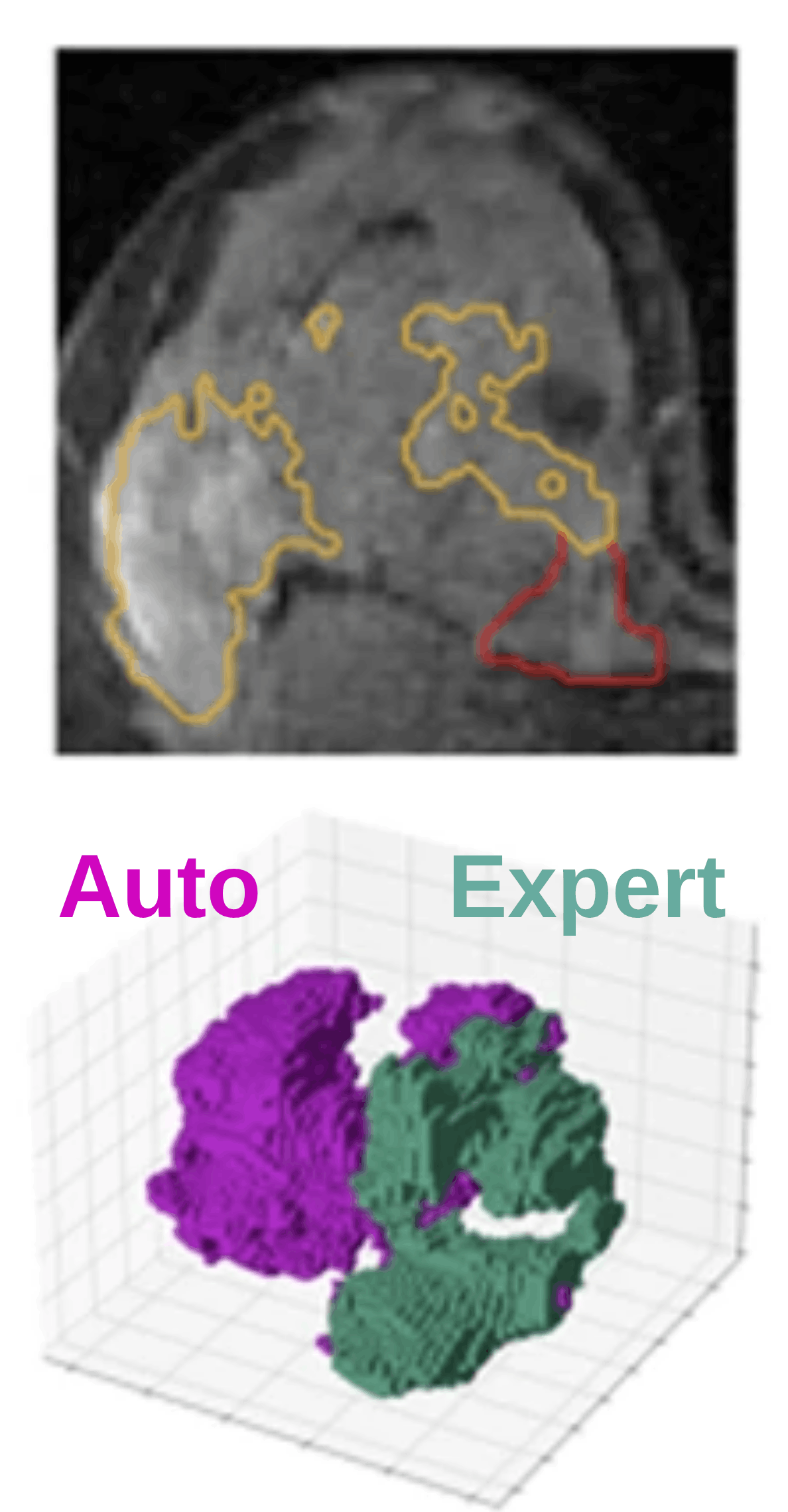}} \tabularnewline

\bottomrule
\end{tabular}
\end{adjustbox}

\label{tab:datasets}
\end{table*}

\subsection{FNLL benchmarking framework for cross-silo medical image segmentation}
\label{sec2_framework}

\textbf{Federated segmentation framework.}
We build the benchmark model on a federated adaptation of nnU-Net \cite{isensee2021nnu, isensee2024nnu}, whose data-driven self-configuration provides a strong and task-adaptive basis for comparing FNLL methods across diverse segmentation datasets.
In the federated setting, local data fingerprints are aggregated into a global fingerprint for unified experiment planning and preprocessing, ensuring consistently configured networks and data processing across clients \cite{kades2022towards, bujotzek2025real}.

\textbf{Client-noise scenarios.}
FNLL robustness is evaluated in four client-noise scenarios: \textit{clean}, where all clients use clean labels; \textit{noisy}, where all clients use noisy labels; \textit{ratio of clients (roc)}, where a fraction $p$ of clients is fully noisy and the remaining clients are clean; and \textit{ratio on all (roa)}, where each client contains a fraction $p$ of noisy samples.

\textbf{Evaluation.}
While training labels are inherently noisy depending on the client-noise scenario, evaluation is performed on clean masks to quantify robustness against label noise with respect to expert or consensus references \cite{wei2024deep}.
General segmentation performance is assessed using the overlap-based Dice score.
To characterize robustness to specific label-noise types, we additionally use noise-type-specific metrics: HD95 for contour disagreement, foreground-background instance-level F1 for missing or additional target instances, and voxel-based class confusion for label confusion.
The F1 score matches instances at an IoU threshold of 0.1, while class confusion measures, for each ground-truth class, the fraction of its voxels predicted as another foreground class (Equation~\ref{eq:clsconf}).
Metric edge-case handling of absent classes and empty masks, is described in Appendix~\ref{app0:metric_edgecases}.

\begin{equation}
\mathrm{ClsConf}_{c_i}
=
\frac{
\left|\left\{ v \in \Omega \;:\; y_v = c_i,\ \hat{y}_v \in \mathcal{C}_{\mathrm{fg}} \setminus \{c_i\} \right\}\right|
}{
\left|\left\{ v \in \Omega \;:\; y_v = c_i \right\}\right|
}
\label{eq:clsconf}
\end{equation}

We perform bootstrap resampling of each evaluation set with 1000 iterations and sample size $N$, then average scores over classes, clients, and folds.
To identify the best-performing FNLL methods, metric scores were averaged across classes, clients, folds, and datasets, then ranked per client-noise scenario and after averaging across all scenarios.
To assess whether FNLL methods improve over FedAvg, we perform one-sided paired Wilcoxon signed-rank tests on case-level score differences matched by case identifier within each dataset and pooled across datasets for each metric, client-noise scenario, and method. Holm-Bonferroni correction is applied across method-wise comparisons against FedAvg to limit false-positive findings.

\subsection{Benchmarked segmentation label noise mitigating methods}
\label{sec2_fnllmethods}
We benchmark representative methods from major FNLL approaches, selected for applicability to real-world segmentation label noise, compatibility with nnU-Net, and feasibility in cross-silo healthcare settings without central expert validation data.
Methods are compared against FedAvg as the default FL baseline \cite{mcmahan2017communication}.

\textbf{Noise-aware aggregation: FedA³I \cite{wu2024feda3i}.}
FedA³I addresses heterogeneous segmentation label noise by estimating client-specific segmentation tendencies for aggregation. After a FedAvg warm-up, a server-side two-component GMM separates over- and under-segmenting clients and enables quality-aware, layer-wise aggregation. We selected FedA³I as it targets general contour-related annotation bias, whereas alternatives focus on incompleteness-only noise \cite{xiang2024fedia} or combine aggregation with binary label correction \cite{zhu2023feddm}. Key hyperparameters are warm-up rounds and the inter-group weighting coefficient.

\textbf{Noise-robust personalization: IOP-FL \cite{jiang2023iop}.}
IOP-FL personalizes federated training by combining local and global gradient information to guide optimization toward client-specific optima. We selected it for its architecture-agnostic design, enabling direct integration into our nnU-Net-based framework. We use the inside-federation variant, which combines current and historical model information through a single mixing factor $\alpha$.

\textbf{Label correction: FedCorr \cite{xu2022fedcorr}.}
FedCorr implements a staged, task-agnostic label-correction pipeline without requiring noise assumptions or clean central validation data \cite{xiang2024fedia, zhu2023feddm, bai2024improving}. After initial pre-processing, it identifies noisy clients via LID statistics, separates clean and noisy samples using a loss-based GMM, and corrects selected labels using global predictions with noise-adaptive proximal regularization. Key hyperparameters control pre-processing rounds, relabeling extent and confidence, and regularization strength.

\textbf{Sample selection: FedSelect \cite{zhang2025proxy}.}
FedSelect mitigates label noise via meta-learned importance-aware sample and client selection, without requiring an external clean validation set. Sample importance is inferred from training dynamics and guides both sample selection and client weighting using client-side proxy validation data. Key hyperparameters control warm-up rounds, selected client and sample fractions, meta-margin momentum, and proxy validation set size.

\subsection{Experimental setup}
\label{sec2_expsetup}

Datasets were partitioned into 3-5 FL clients based on clinically or technically meaningful factors, including center, scanner, sub-dataset, rater identity, or animal assignment (Table~\ref{tab:expsetup_merged}).

The dataset-specific client composition determines the partially noisy client-noise scenarios \textit{roa} and \textit{roc}.
In \textit{roa}, a nominal noise level of $p=50\%$ is applied within each client.
In \textit{roc}, $p$ specifies the proportion of noisy clients, with $|K_{\mathrm{noisy}}| = \lceil p\,|K_{\mathrm{all}}| \rceil$. Since client counts and sample counts per client vary across datasets, we report the effective noise level $p_{\mathrm{eff}}$ as the fraction of noisy samples:
\[
p_{\mathrm{eff}}^{\textit{roa}}(p)
=
\frac{\sum_{k \in K_{\mathrm{all}}} n_k\,p}{\sum_{k \in K_{\mathrm{all}}} n_k}
\approx p,
\qquad
p_{\mathrm{eff}}^{\textit{roc}}
=
\frac{\sum_{k \in K_{\mathrm{noisy}}} n_k}{\sum_{k \in K_{\mathrm{all}}} n_k}.
\]
Here, $K_{\mathrm{all}} = K_{\mathrm{clean}} \cup K_{\mathrm{noisy}}$ and $n_k$ denotes the number of samples at client $k$. Corresponding sample counts and resulting effective noise levels $p_{\mathrm{eff}}$ are listed in Table~\ref{tab:expsetup_merged}.

\begin{table*}[t]
\centering

\caption{Overview of dataset-specific federated client partitioning and the resulting partially noisy client-noise scenarios. For \textit{roa} and \textit{roc}, sample counts are reported as $n_{\mathrm{clean}}/n_{\mathrm{noisy}}$. In \textit{roc}, $C$ and $N$ denote the sets of clean and noisy clients, respectively.}

\footnotesize
\setlength{\tabcolsep}{4pt}
\renewcommand{\arraystretch}{1.12}

\begin{adjustbox}{max width=\textwidth}
\begin{tabular}{
    M{0.9cm}
    L{2.15cm}
    *{6}{L{2.05cm}}
}
\toprule
& &
\multicolumn{1}{c}{\textbf{LIDC}} &
\multicolumn{1}{c}{\textbf{RIGA}} &
\multicolumn{1}{c}{\textbf{GleasonHD}} &
\multicolumn{1}{c}{\textbf{MouseT}} &
\multicolumn{1}{c}{\textbf{MMIS}} &
\multicolumn{1}{c}{\textbf{MMIA}} \\
\midrule

\multirowcell{3}[-3ex][c]{\rotatebox[origin=c]{90}{\textbf{FL splitting}}}
& \textit{\# clients}
& \centering 4
& \centering 3
& \centering 3
& \centering 5
& \centering 4
& \centering 4 \tabularnewline

& \textit{Split criterion}
& \makecell[tl]{scanner\\manufacturer}
& \makecell[tl]{sub-dataset\\membership}
& \makecell[tl]{random}
& \makecell[tl]{random\\mouse}
& \makecell[tl]{rater\\identity}
& \makecell[tl]{originating\\center}\tabularnewline

& \textit{Samples per client}
& \makecell[tl]{C0: 1281\\C1: 218\\C2: 475\\C3: 147}
& \makecell[tl]{C0: 195\\C1: 94\\C2: 460}
& \makecell[tl]{C0: 159\\C1: 158\\C2: 159}
& \makecell[tl]{C0: 178\\C1: 127\\C2: 67\\C3: 48\\C4: 32}
& \makecell[tl]{C0: 34\\C1: 29\\C2: 27\\C3: 30}
& \makecell[tl]{C0: 291\\C1: 171\\C2: 980\\C3: 64}\tabularnewline

\midrule

\multirowcell{5}[-1ex][c]{\rotatebox[origin=c]{90}{\textbf{Partial noise}}}
& \textit{roa} ($n_{\mathrm{c}}/n_{\mathrm{n}}$)
& \centering 1062 / 1059
& \centering 375 / 374
& \centering 239 / 237
& \centering 227 / 225
& \centering 61 / 59
& \centering 754 / 752 \tabularnewline

& $p_{\mathrm{eff}}^{\textit{roa}}$
& \centering 49.93\%
& \centering 49.93\%
& \centering 49.79\%
& \centering 49.78\%
& \centering 49.17\%
& \centering 49.93\% \tabularnewline

\cmidrule(lr){2-8}


& \textit{roc} clients
& \makecell[tl]{$C=\{0,1\}$\\$N=\{2,3\}$}
& \makecell[tl]{$C=\{0\}$\\$N=\{1,2\}$}
& \makecell[tl]{$C=\{0\}$\\$N=\{1,2\}$}
& \makecell[tl]{$C=\{0,1,2\}$\\$N=\{3,4\}$}
& \makecell[tl]{$C=\{0,1\}$\\$N=\{2,3\}$} 
& \makecell[tl]{$C=\{0,1\}$\\$N=\{2,3\}$} \tabularnewline

& \textit{roc} ($n_{\mathrm{c}}/n_{\mathrm{n}}$)
& \centering 1499 / 622
& \centering 195 / 554
& \centering 159 / 317
& \centering 305 / 147
& \centering 63 / 57 
& \centering 462 / 1044 \tabularnewline

& $p_{\mathrm{eff}}^{\textit{roc}}$
& \centering 29.33\%
& \centering 73.97\%
& \centering 66.60\%
& \centering 32.52\%
& \centering 47.50\%
& \centering 69.32\% \tabularnewline

\bottomrule
\end{tabular}
\end{adjustbox}

\label{tab:expsetup_merged}
\end{table*}

The federated framework uses the default nnU-Net full-resolution ResEncM configuration with dataset-specific patch and batch sizes (Appendix~\ref{app1:data_hparams}).
For each client configuration, communication rounds were set to match cumulatively the nnU-Net recommendation of 1000 local epochs \cite{isensee2021nnu}, using one local epoch per round.
Standard nnU-Net five-fold splits were generated per client; experiments used the first three folds, with metrics computed on the corresponding held-out validation sets.

Prior to benchmarking, method-specific FNLL hyperparameters were optimized by grid search over predefined candidate configurations, with final settings selected by average performance across datasets. Details are provided in Appendix~\ref{app1:fnll_hparams}.

\section{Results}\label{sec:results}

\subsection{In-depth data and label noise analysis}
\begin{figure}[h!]
\centering
\includegraphics[width=\textwidth]{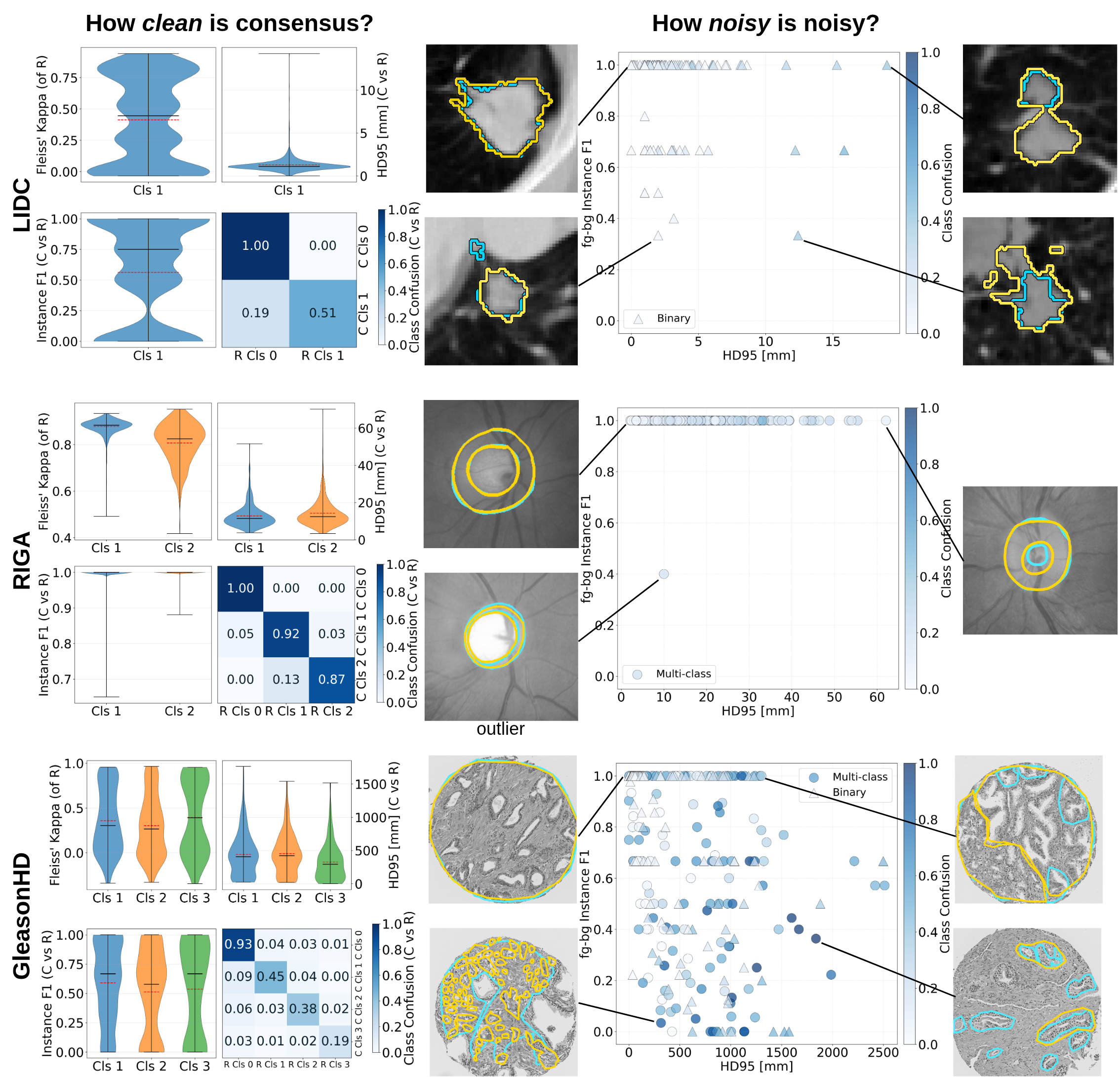}
\caption{Consensus- and noisy label characteristics of LIDC, RIGA and GleasonHD datasets including representative examples illustrating characteristic label noise patterns. In left column, R and C abbreviate raters and consensus; in right column, clean/consensus contours in yellow and noisy contours in cyan.}
\label{fig:data_analysis_lidcrigagleason}
\end{figure}

\begin{figure}[h!]
\centering
\includegraphics[width=\textwidth]{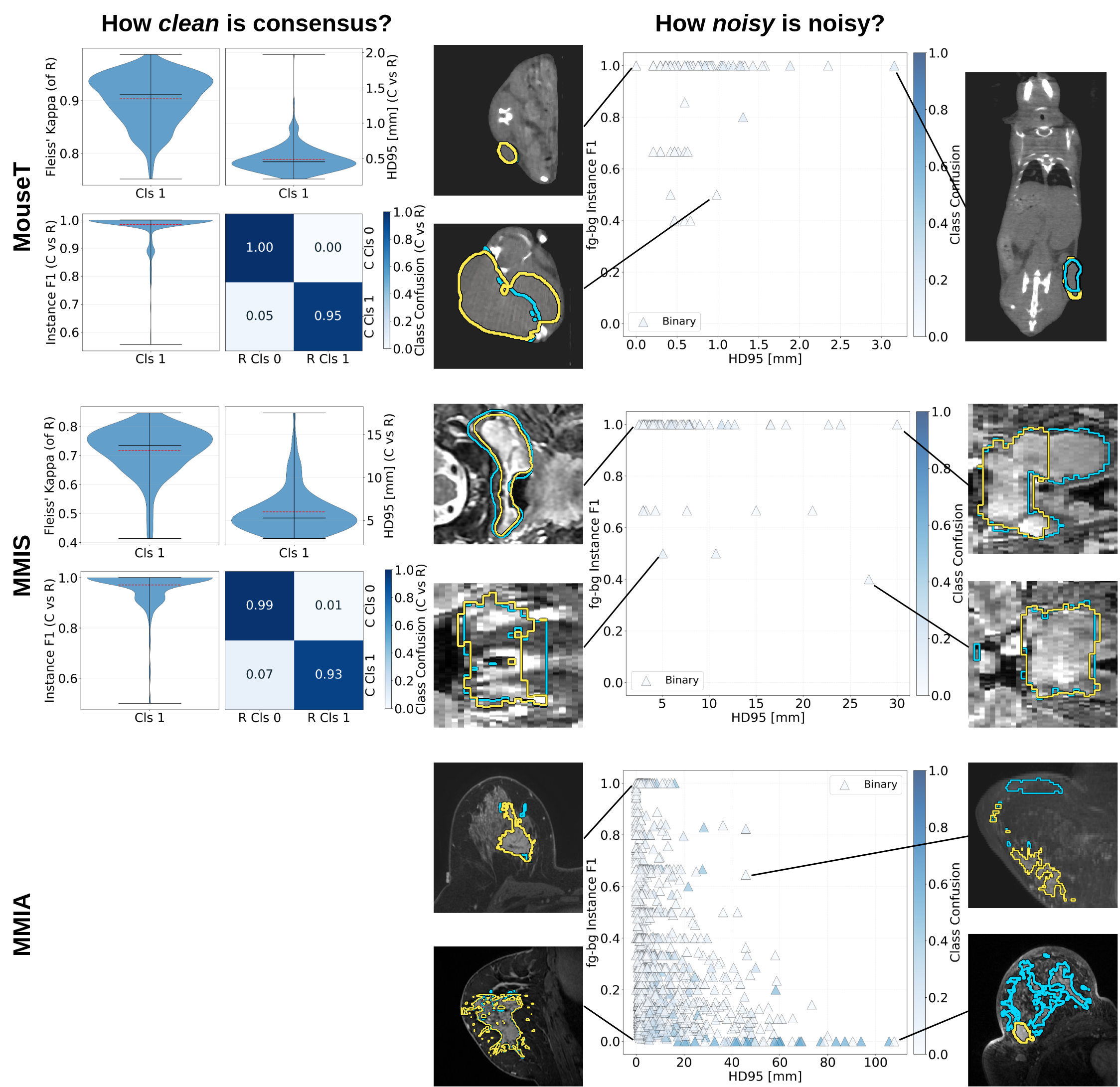}
\caption{Consensus- and noisy label characteristics of MouseT, MMIS and MMIA datasets including representative examples illustrating characteristic label noise patterns (yellow=consensus/expert, cyan=noise). As MMIA comes out-of-the-box with expert (clean) and automatically generated (noisy) label masks, it is not included in the \textit{How clean is consensus} considerations.}
\label{fig:data_analysis_mousetmmismmia}
\end{figure}

Figures~\ref{fig:data_analysis_lidcrigagleason} and \ref{fig:data_analysis_mousetmmismmia} characterize the six inherently noisy benchmark datasets by consensus-mask quality and noisy-mask properties. 
Consensus cleanliness is assessed using inter-rater and rater-consensus agreement via class-wise Fleiss' kappa, HD95, instance-level F1, and class confusion.
Noisy masks are analyzed in the joint space of these noise-sensitive metrics to quantify noise-type prevalence and severity, complemented by representative examples from regions enriched for characteristic noise patterns.

\textbf{How \textit{clean} is consensus?}
Consensus quality varies substantially across datasets.
\textit{RIGA} and \textit{MouseT} show the cleanest consensus masks, with high inter-rater and rater-consensus agreement across metrics.
\textit{LIDC} and \textit{MMIS} show intermediate consensus quality, with increased variability mainly in boundary consistency reflected by HD95.
\textit{GleasonHD} has the lowest consensus cleanliness, with reduced Fleiss' kappa, weaker rater-consensus agreement, and more pronounced off-diagonal class-confusion entries.
\textit{MMIA} is excluded because clean labels are provided as expert annotations rather than multi-rater consensus.

\textbf{How \textit{noisy} is noisy?}
\textit{RIGA} shows the simplest noise regime, with high instance-level agreement, negligible class confusion, and variable HD95, indicating predominantly contour-driven noise.
\textit{LIDC}, \textit{MouseT}, and \textit{MMIS} show mixed noise, dominated by contour disagreement but with reduced instance-level F1 indicating missed or additional target structures.
\textit{MMIA} is mainly characterized by missed or additional target structures, reflected by shifts along the instance-level F1 axis.
\textit{GleasonHD} shows the most severe and heterogeneous label noise, with broad dispersion across all three metric dimensions.

\subsection{Comparative benchmarking of FNLL methods}

\begin{figure}[h!]
\centering
\includegraphics[width=\textwidth]{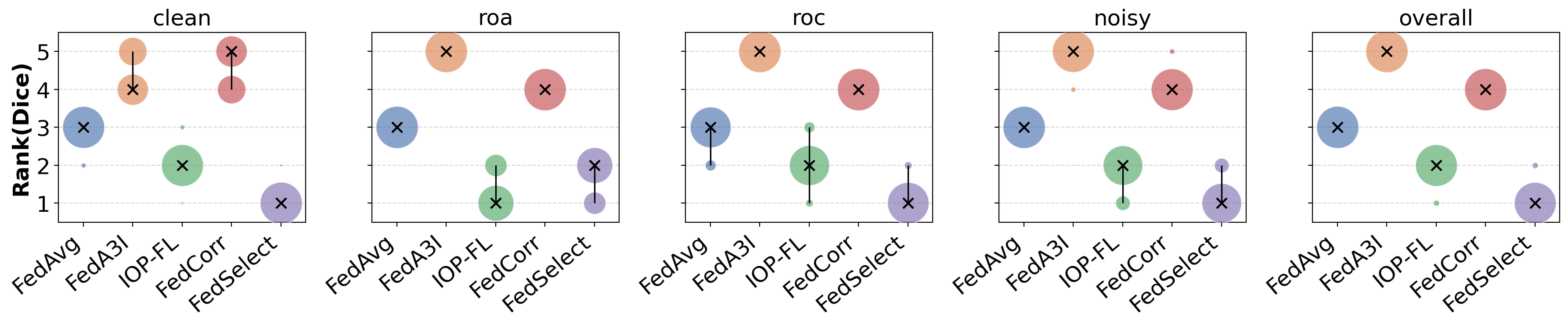}
\caption{Rank stability of Dice segmentation performances based on bootstrapping across datasets, client-noise scenarios and methods. Lower ranks indicate better performances, bubble size reflects the ranking frequency across datasets.}

\label{fig:seg_perf_dice_rank}
\end{figure}

\textbf{Which FNLL method performs best across datasets, noise types and scenarios?}
Across datasets and client-noise scenarios, \textit{FedSelect} is the strongest overall FNLL method due to its high rank stability and consistent performance, whereas \textit{IOP-FL} is the main competitor and achieves the best Dice scores in many individual dataset-scenario comparisons (Table~\ref{tab:bootstrap_dice_results}, Figure~\ref{fig:seg_perf_dice_rank}, Appendix~\ref{app2:general_dice}).
\textit{FedAvg} remains a strong baseline, while \textit{FedCorr} shows dataset-dependent gains and \textit{FedA3I} performs weakest overall, both frequently failing to improve over \textit{FedAvg}.
Statistical testing confirms this mixed picture: improvements over \textit{FedAvg} are only partially significant, with corrected significant Dice gains observed primarily for \textit{IOP-FL} in the \textit{roa}, \textit{roc}, and \textit{noisy} scenarios (Appendix~\ref{app3:statistics}).
Segmentation performance is strongly shaped by dataset difficulty and label-noise characteristics (Figures~\ref{fig:data_analysis_lidcrigagleason} and \ref{fig:data_analysis_mousetmmismmia}).
\textit{RIGA} and \textit{MouseT} are comparatively easy, contour-dominated datasets with high Dice scores and limited separation between top methods, although \textit{FedSelect} and \textit{IOP-FL} remain strongest overall.
\textit{GleasonHD}, which combines all three noise types, is the most challenging dataset, with substantially lower performance, broader distributions, and strongest results for \textit{FedCorr}.
\textit{LIDC}, \textit{MMIS}, and \textit{MMIA} form an intermediate regime; \textit{FedSelect} and \textit{IOP-FL} perform similarly under contour-related noise, while \textit{IOP-FL} performs best on the predominantly instance-level noise in \textit{MMIA}.
The ranking analysis (Figure~\ref{fig:seg_perf_dice_rank}) confirms a stable overall ordering, with \textit{FedSelect} ranked first, followed by \textit{IOP-FL}; \textit{FedAvg} remains competitive and outperforms the remaining FNLL methods. This identifies \textit{FedSelect} and \textit{IOP-FL} as the top-performing FNLL tier across noisy and even beneficial in clean scenarios.

\begin{table*}[h!]
\centering

\caption{Mean validation Dice values ($\times 100$) for each dataset, noise scenario, and method. Dice scores $\times 100$; best values in bold and green, second best underlined, worst in red.}

\small
\setlength{\aboverulesep}{0.5pt}
\setlength{\belowrulesep}{0.5pt}
\setlength{\cmidrulesep}{0.5pt}

\begin{tabular}{llccccc}
\toprule
\textit{Dataset} & \textit{Scenario} & \textbf{FedAvg} & \textbf{FedA3I} & \textbf{IOP-FL} & \textbf{FedCorr} & \textbf{FedSelect} \\
\midrule

\multirow{4}{*}{LIDC} & clean & \cellcolor[HTML]{E3F3B6}$56.4$ & \cellcolor[HTML]{C9667D}$52.9$ & \cellcolor[HTML]{66A487}$\mathbf{58.6}$ & \cellcolor[HTML]{FCC099}$54.4$ & \cellcolor[HTML]{BDE3A4}$\underline{57.1}$ \\
& roa & \cellcolor[HTML]{EDF7C0}$56.7$ & \cellcolor[HTML]{C9667D}$54.1$ & \cellcolor[HTML]{66A487}$\mathbf{58.5}$ & \cellcolor[HTML]{DE7A7D}$54.4$ & \cellcolor[HTML]{A9DAA2}$\underline{57.6}$ \\
& roc & \cellcolor[HTML]{E9F6BB}$52.6$ & \cellcolor[HTML]{F9A98F}$51.6$ & \cellcolor[HTML]{66A487}$\mathbf{53.6}$ & \cellcolor[HTML]{C9667D}$51.1$ & \cellcolor[HTML]{A1D7A1}$\underline{53.1}$ \\
& noisy & \cellcolor[HTML]{C4E6A5}$49.5$ & \cellcolor[HTML]{E9F6BB}$49.1$ & \cellcolor[HTML]{66A487}$\mathbf{50.5}$ & \cellcolor[HTML]{C9667D}$46.9$ & \cellcolor[HTML]{72BA92}$\underline{50.3}$ \\
\cmidrule(lr){1-7}

\multirow{4}{*}{RIGA} & clean & \cellcolor[HTML]{67A688}$\underline{93.2}$ & \cellcolor[HTML]{FAB192}$84.7$ & \cellcolor[HTML]{66A487}$\mathbf{93.2}$ & \cellcolor[HTML]{C9667D}$82.2$ & \cellcolor[HTML]{68A889}$93.1$ \\
& roa & \cellcolor[HTML]{66A487}$\mathbf{92.8}$ & \cellcolor[HTML]{F8A68E}$81.5$ & \cellcolor[HTML]{68A889}$\underline{92.6}$ & \cellcolor[HTML]{C9667D}$78.8$ & \cellcolor[HTML]{6AAB8B}$92.4$ \\
& roc & \cellcolor[HTML]{66A487}$\mathbf{89.0}$ & \cellcolor[HTML]{DF7B7D}$80.3$ & \cellcolor[HTML]{67A688}$\underline{88.9}$ & \cellcolor[HTML]{C9667D}$79.5$ & \cellcolor[HTML]{68A989}$88.8$ \\
& noisy & \cellcolor[HTML]{6BAD8C}$86.1$ & \cellcolor[HTML]{FBB896}$77.3$ & \cellcolor[HTML]{66A487}$\mathbf{86.5}$ & \cellcolor[HTML]{C9667D}$74.4$ & \cellcolor[HTML]{69AA8A}$\underline{86.2}$ \\
\cmidrule(lr){1-7}

\multirow{4}{*}{GleasonHD} & clean & \cellcolor[HTML]{FEDCAC}$32.1$ & \cellcolor[HTML]{C9667D}$28.4$ & \cellcolor[HTML]{FFF6C9}$33.2$ & \cellcolor[HTML]{66A487}$\mathbf{39.1}$ & \cellcolor[HTML]{8ECD9C}$\underline{37.4}$ \\
& roa & \cellcolor[HTML]{FEEAB7}$33.5$ & \cellcolor[HTML]{C9667D}$30.6$ & \cellcolor[HTML]{FFFCD3}$\underline{34.2}$ & \cellcolor[HTML]{66A487}$\mathbf{38.1}$ & \cellcolor[HTML]{FED5A6}$33.0$ \\
& roc & \cellcolor[HTML]{DDF1B2}$\underline{32.4}$ & \cellcolor[HTML]{E9867F}$29.4$ & \cellcolor[HTML]{C9667D}$28.7$ & \cellcolor[HTML]{66A487}$\mathbf{34.5}$ & \cellcolor[HTML]{EDF7C0}$32.1$ \\
& noisy & \cellcolor[HTML]{FFF4C6}$30.6$ & \cellcolor[HTML]{FFF0C0}$30.4$ & \cellcolor[HTML]{C9667D}$27.1$ & \cellcolor[HTML]{66A487}$\mathbf{35.0}$ & \cellcolor[HTML]{FFFBD2}$\underline{30.9}$ \\
\cmidrule(lr){1-7}

\multirow{4}{*}{MouseT} & clean & \cellcolor[HTML]{66A487}$\mathbf{92.5}$ & \cellcolor[HTML]{C9E8A6}$89.9$ & \cellcolor[HTML]{98D29F}$90.9$ & \cellcolor[HTML]{C9667D}$83.6$ & \cellcolor[HTML]{67A688}$\underline{92.5}$ \\
& roa & \cellcolor[HTML]{66A487}$\mathbf{92.7}$ & \cellcolor[HTML]{C9667D}$87.7$ & \cellcolor[HTML]{AFDDA3}$91.5$ & \cellcolor[HTML]{FFF3C4}$89.9$ & \cellcolor[HTML]{87C99A}$\underline{92.0}$ \\
& roc & \cellcolor[HTML]{6DB18D}$91.0$ & \cellcolor[HTML]{C9667D}$83.1$ & \cellcolor[HTML]{68A889}$\underline{91.3}$ & \cellcolor[HTML]{F1F9C6}$87.8$ & \cellcolor[HTML]{66A487}$\mathbf{91.4}$ \\
& noisy & \cellcolor[HTML]{6CAF8C}$\underline{90.5}$ & \cellcolor[HTML]{C9667D}$86.1$ & \cellcolor[HTML]{9ED5A0}$89.8$ & \cellcolor[HTML]{FFFBD2}$88.3$ & \cellcolor[HTML]{66A487}$\mathbf{90.7}$ \\
\cmidrule(lr){1-7}

\multirow{4}{*}{MMIS} & clean & \cellcolor[HTML]{DB777D}$81.2$ & \cellcolor[HTML]{C9667D}$81.0$ & \cellcolor[HTML]{FAB494}$\underline{81.8}$ & \cellcolor[HTML]{E37F7D}$81.3$ & \cellcolor[HTML]{66A487}$\mathbf{84.3}$ \\
& roa & \cellcolor[HTML]{FBBB97}$80.5$ & \cellcolor[HTML]{C9667D}$79.6$ & \cellcolor[HTML]{E1F2B5}$\underline{81.9}$ & \cellcolor[HTML]{FED1A3}$80.7$ & \cellcolor[HTML]{66A487}$\mathbf{83.2}$ \\
& roc & \cellcolor[HTML]{EB8C82}$77.4$ & \cellcolor[HTML]{C9667D}$77.1$ & \cellcolor[HTML]{F0F9C4}$\underline{78.6}$ & \cellcolor[HTML]{E6827D}$77.3$ & \cellcolor[HTML]{66A487}$\mathbf{79.8}$ \\
& noisy & \cellcolor[HTML]{D6727D}$74.8$ & \cellcolor[HTML]{C9667D}$74.6$ & \cellcolor[HTML]{66A487}$\mathbf{77.6}$ & \cellcolor[HTML]{EF9385}$75.1$ & \cellcolor[HTML]{C1E5A5}$\underline{76.7}$ \\
\cmidrule(lr){1-7}

\multirow{4}{*}{MMIA} & clean & \cellcolor[HTML]{93D09D}$\underline{68.3}$ & \cellcolor[HTML]{FFF4C7}$64.8$ & \cellcolor[HTML]{66A487}$\mathbf{69.8}$ & \cellcolor[HTML]{C9667D}$60.8$ & \cellcolor[HTML]{A1D7A1}$68.0$ \\
& roa & \cellcolor[HTML]{C9667D}$61.5$ & \cellcolor[HTML]{A9DAA2}$65.5$ & \cellcolor[HTML]{66A487}$\mathbf{66.6}$ & \cellcolor[HTML]{97D19E}$65.7$ & \cellcolor[HTML]{74BE95}$\underline{66.1}$ \\
& roc & \cellcolor[HTML]{C9667D}$63.3$ & \cellcolor[HTML]{91CF9D}$66.4$ & \cellcolor[HTML]{66A487}$\mathbf{67.1}$ & \cellcolor[HTML]{66A487}$\underline{67.1}$ & \cellcolor[HTML]{95D09E}$66.4$ \\
& noisy & \cellcolor[HTML]{C9667D}$63.1$ & \cellcolor[HTML]{E7847E}$63.8$ & \cellcolor[HTML]{66A487}$\mathbf{69.3}$ & \cellcolor[HTML]{FEE4B2}$65.4$ & \cellcolor[HTML]{B4DFA3}$\underline{67.8}$ \\

\bottomrule

\end{tabular}

\label{tab:bootstrap_dice_results}
\end{table*}

\begin{figure}[h!]
\centering
\includegraphics[width=\textwidth]{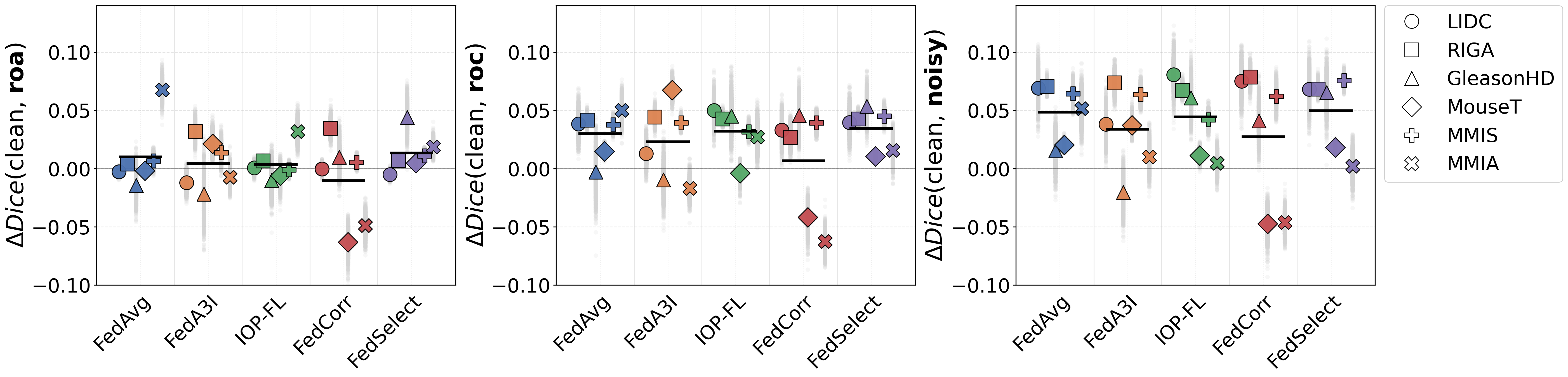}
\caption{Clean-referenced robustness of FNLL methods across the client-noise scenarios \textit{roa}, \textit{roc}, and \textit{noisy}, shown as the absolute Dice difference to clean performance. Black bars denote the mean across datasets, and light gray points the corresponding bootstrap distribution.}
\label{fig:robustness_analysis}
\end{figure}

\textbf{How robust are FNLL methods to partially and fully noisy training?}
Clean-referenced Dice drops show that all noisy regimes remain close to clean baselines, but differ by scenario and method (Figure~\ref{fig:robustness_analysis}).
The \textit{roa} setting, with $\sim$50\% noisy samples per client, induces only minor degradation, indicating robustness to within-client partial noise. In contrast, \textit{roc} (29.33-73.97\% effective noise; mean 53.2\%, Table~\ref{tab:expsetup_merged}) causes larger losses and is more harmful than \textit{roa}, while fully noisy training yields the strongest degradation.
Method-wise, \textit{IOP-FL} remains closest to the clean baseline in \textit{roa}, whereas \textit{FedSelect} shows the largest drop. In \textit{roc} and fully noisy settings, \textit{FedCorr} achieves the smallest Dice degradation, while \textit{FedSelect} again degrades most. This contrasts with absolute performance (Table~\ref{tab:bootstrap_dice_results}, Figure~\ref{fig:seg_perf_dice_rank}), where \textit{FedSelect} ranks best overall, indicating that highest Dice does not coincide with minimal degradation and that robustness is scenario- and dataset-dependent.

\textbf{Which FNLL method mitigates specific label noise types best?}
We assess noise-type-specific robustness using dedicated metrics (Section~\ref{sec2_framework}), including for each metric only datasets where the corresponding noise type is present according to Figures~\ref{fig:data_analysis_lidcrigagleason} and \ref{fig:data_analysis_mousetmmismmia}. This preserves noise-type specificity while retaining realistic mixed-noise settings.

\begin{figure}[h!]
\centering
\includegraphics[width=\textwidth]{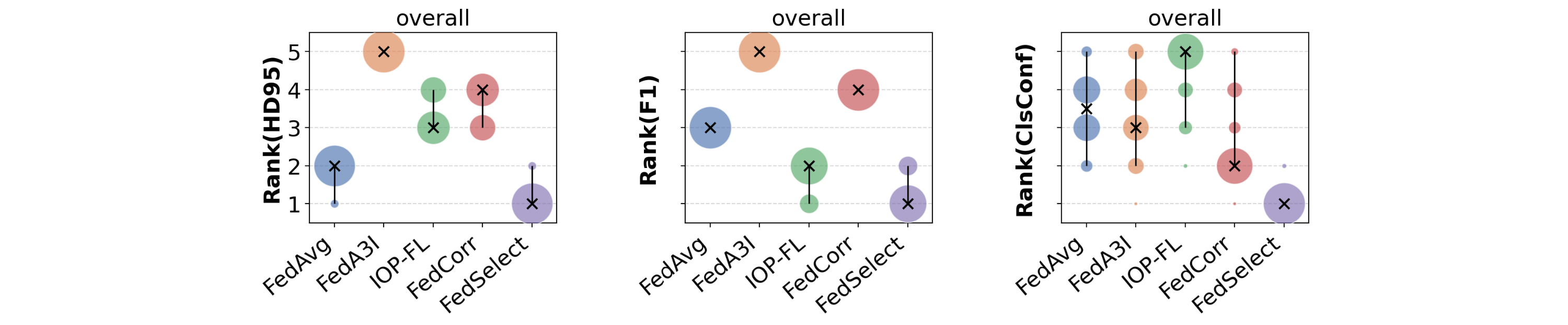}
\caption{Rank stability of segmentation performances w.r.t. noise-type-specific metrics based on bootstrapping across datasets, client-noise scenarios and methods. Lower ranks indicate better performance, bubble size reflects the ranking frequency across datasets.}
\label{fig:method_noisetype_mitigation}
\end{figure}

For contour disagreement, assessed by HD95, \textit{FedSelect} ranks best overall and is the only FNLL method consistently matching or outperforming the strong \textit{FedAvg} baseline (Figure~\ref{fig:method_noisetype_mitigation}, Appendix~\ref{app2:noisespecific_hd95}). Separation is clearest in contour-dominated datasets (\textit{RIGA}, \textit{MouseT}, \textit{MMIS}), where \textit{FedSelect} yields the lowest boundary deviations across scenarios, while \textit{FedA3I} and \textit{FedCorr} often show substantially higher HD95.
For missing or additional target structures, assessed by instance-wise F1, \textit{FedSelect} again ranks first overall, with \textit{IOP-FL} and \textit{FedAvg} as strongest competitors. Separation is clearest in the instance-noise-dominated \textit{MMIA}, where \textit{FedSelect} and \textit{IOP-FL} achieve the highest instance-wise agreement (Appendix~\ref{app2:noisespecific_f1}).
For label swapping, assessed by class confusion, only \textit{GleasonHD} is included. \textit{FedSelect} ranks best overall, followed by \textit{FedCorr} (Appendix~\ref{app2:noisespecific_clsconf}). Given the single dataset, mixed noise, low performance, and limited consensus due to strong inter-rater disagreement, conclusions should be interpreted with caution.
Statistical testing supports these trends selectively: significant improvements over \textit{FedAvg} are observed primarily for \textit{FedSelect} in instance-level robustness (F1) across scenarios and for both \textit{FedSelect} and \textit{IOP-FL} in contour-related settings (HD95, \textit{roc}), while no significant gains are found for class-confusion robustness (Appendix~\ref{app3:statistics}).

Overall, the noise-type-specific analysis identifies \textit{FedSelect} as the most consistent method across all three noise dimensions, supporting the Dice-based results and confirming it as the strongest FNLL method in our benchmark.
We translate the combined Dice and noise-sensitive ranking analyses into a decision guide (Table~\ref{tab:fnll_decision_guide}; Appendix~\ref{app2:general_dice}, \ref{app2:noisespecific_hd95}, \ref{app2:noisespecific_f1}, \ref{app2:noisespecific_clsconf}) to support targeted FNLL method selection in deployment-relevant noisy federated settings.
Recommended methods are based on lowest mean bootstrap rank, while bold entries indicate cases with statistically significant improvements over FedAvg after Holm-Bonferroni correction.

\begin{table}[t]
\centering
\caption{Decision guide for selecting FNLL methods according to the dominant segmentation label-noise type and client-noise scenario.}
\label{tab:fnll_decision_guide}
\begin{tabular}{lccccc}
\toprule
\textbf{Noise type (metric)} & \multicolumn{5}{c}{\textbf{Client-noise scenario}} \\
\cmidrule(lr){2-6}
 & \textbf{clean} & \textbf{\textit{roa}} & \textbf{\textit{roc}} & \textbf{noisy} & \textbf{overall} \\
\midrule
\textbf{Contour} (HD95) & FedSelect & FedAvg & \textbf{FedSelect} & FedAvg & FedSelect \\
\textbf{Instance} (F1) & \textbf{FedSelect} & IOP-FL & \textbf{FedSelect} & \textbf{FedSelect} & \textbf{FedSelect} \\
\textbf{Confusion} (ClsConf) & FedSelect & FedCorr & FedAvg & FedSelect & FedSelect \\
\midrule
\textbf{General} (Dice) & FedSelect & \textbf{IOP-FL} & FedSelect & FedSelect & FedSelect \\
\bottomrule
\end{tabular}
\vspace{0.35em}
\begin{minipage}{0.98\linewidth}
\footnotesize Recommended methods are selected by the lowest mean bootstrap rank for each metric and scenario. Bold entries indicate Holm-Bonferroni corrected significance versus FedAvg.
\end{minipage}
\end{table}

\section{Discussion}\label{sec:discussion}
This study addresses a central gap in FNLL for medical image segmentation: the lack of a standardized benchmark that combines \textit{diverse datasets}, \textit{numerous real-world} segmentation label noise types, \textit{multiple client-noise scenarios}, and task-relevant evaluation to enable informed decision-making for FNLL method selection.

The included datasets span diverse modalities, dimensionalities, target structures, and noise origins, and the in-depth analysis shows that real-world segmentation label noise occurs both in isolation and in combinations of contour disagreement, missing or additional target structures, and class confusion.
Noise-sensitive metrics enable explicit noise-type characterization, supporting both dataset interpretation and analysis of method behavior across noise regimes.
Consensus quality varies across datasets: \textit{GleasonHD} shows the lowest consensus cleanliness and most severe heterogeneous noisy-mask characteristics, partly explaining its role as the most challenging benchmark case and warranting caution for label-confusing conclusions.
Overall, however, the consensus analysis indicates that, except for \textit{GleasonHD}, evaluation against the derived clean reference provides a fair basis for method comparison.

The comparative benchmark shows that the combination of realistic data with noise and standardized evaluation is sufficiently discriminative to reveal meaningful differences between FNLL strategies. Across datasets and client-noise scenarios, \textit{FedSelect} emerges as the strongest overall method, with \textit{IOP-FL} as the only comparable alternative, while \textit{FedAvg} remains a strong baseline and \textit{FedA3I} and \textit{FedCorr} are less consistently competitive. Importantly, dedicated FNLL methods improve over the widely used \textit{FedAvg} baseline only in specific noise regimes, highlighting the need for informed method selection rather than assuming universal FNLL superiority. This is further reflected in the statistical analysis, where only a subset of these improvements are significant, indicating that performance gains over \textit{FedAvg} are not consistently robust across all settings.
At the same time, ``best'' depends on the notion of robustness considered. \textit{FedSelect} achieves the highest absolute performance and is most consistent across noise-type-specific analyses, whereas \textit{FedCorr} shows the smallest clean-referenced Dice degradation in the more harmful \textit{roc} and fully noisy settings, largely driven by its performance on the challenging \textit{GleasonHD} dataset. This distinction between absolute performance and clean-referenced robustness would remain hidden in narrower benchmarks.
To facilitate practical use, we summarize the relationships between label-noise characteristics, client-noise scenarios, and FNLL performance in an actionable decision guide (Table~\ref{tab:fnll_decision_guide}), combining rank-based recommendations with statistical evidence against \textit{FedAvg}. The guide identifies \textit{FedSelect} as a robust default and \textit{FedAvg} or \textit{IOP-FL} as preferable in specific settings, supporting informed FNLL method selection in deployment-relevant federated settings.

Although the benchmark suite combines curated datasets, representative FNLL methods, and a comprehensive evaluation protocol, several limitations remain.
The results represent a structured snapshot of the current FNLL landscape, with hyperparameters optimized for average performance across datasets, favoring comparability over dataset-specific optimality.
Comparisons between partially noisy settings (\textit{roa} vs. \textit{roc}) should be interpreted with care, as effective noise levels vary substantially in \textit{roc}, making both settings complementary rather than directly comparable robustness probes.
Noise representation is imbalanced: contour-related noise dominates, instance-level noise is mainly represented by \textit{MMIA}, and class-confusion noise primarily by \textit{GleasonHD}. Accordingly, conclusions are strongest for contour robustness, moderate for instance-level noise, and preliminary for class-swapping.
The decision guide should be interpreted as a data-driven heuristic, as optimal method selection may vary with dataset-specific characteristics.
Future work should extend the suite with additional datasets featuring instance-level and multiclass confusion noise, and evaluate partial-noise settings across multiple noise levels.

By combining real-world noisy datasets, explicit noise characterization, standardized federated training, and competitive FNLL baselines, the suite supports practical use beyond this comparative study, including FNLL method selection, characterization of new datasets, benchmarking on new data, and reproducible evaluation of future methods under deployment-relevant noisy conditions.
Its main value therefore lies not only in the benchmark conclusions, but in providing a reusable foundation for FNLL method development, fair comparison, and dataset-driven analysis in realistic federated segmentation settings.

\section{Conclusion}\label{sec:conclusion}
We presented a benchmark suite for FNLL in cross-silo medical image segmentation, combining diverse real-world noisy datasets, clinically relevant client-noise scenarios, and targeted evaluation.
The dataset analysis shows that the suite covers multiple practically relevant segmentation label-noise types and provides context for interpreting method behavior across datasets and noise regimes.
The comparative benchmark identifies \textit{FedSelect} as the strongest overall FNLL method and \textit{IOP-FL} as the most competitive alternative, while showing that gains over the strong \textit{FedAvg} baseline remain dataset-, noise-, and scenario-dependent. 
Through an actionable decision guide, the suite supports informed FNLL method selection in deployment-relevant federated settings.
Beyond the present benchmark, the released suite provides a reusable foundation for fair comparison, label-noise characterization, and future method development, supporting more reliable federated segmentation and downstream clinical decision-making.

\backmatter

\bmhead{Supplementary information}

Supplementary materials are provided in the Appendix below.

\bmhead{Acknowledgements}
Maximilian Zenk and Ünal Akünal for guidiance in the early phase of the project.

\clearpage

\section*{Statements and Declarations}

\begin{itemize}
\item \textbf{Funding:} This research was funded by the German Federal Ministry of
Education and Research (BMBF) as part of the University Medicine
Network (Project RACOON, 01KX2021), as part of the PrivateAIM project (01ZZ2316M), and as part of the Research Campus M2OLIE, within the Framework “Forschungscampus: Public-private partnership for Innovations” (13GW0388A).

\item \textbf{Competing interests:} The authors have no competing interests to declare that are relevant to the content of this article.

\item \textbf{Author contributions:} Markus Ralf Bujotzek: Conceptualization, Data curation, Formal analysis, Investigation, Methodology, Project administration, Resources, Software, Validation, Visualization, Writing – original draft, Writing – review \& editing. Dimitrios Bounias: Conceptualization, Formal analysis, Investigation, Validation, Writing – review \& editing. Stefan Denner: Conceptualization, Formal analysis, Investigation, Methodology, Writing – review \& editing. Ralf Floca: Conceptualization, Project administration, Supervision, Writing – review \& editing. Maximilian Fischer: Writing – review \& editing. Peter Neher: Supervision, Writing – review \& editing. Klaus H. Maier-Hein: Funding acquisition, Project administration, Resources, Writing – review \& editing.

\item \textbf{Ethics approval:} This study exclusively used publicly available datasets. All data were collected and made available by the original studies in accordance with relevant ethical guidelines and approvals. No additional ethical approval was required for this work.

\item \textbf{Consent to participate:} Not applicable. This study used only publicly available, de-identified data.

\item \textbf{Consent to publish:} Not applicable. This study did not involve any identifiable individual data.

\item \textbf{Data availability:} All datasets used in this study are publicly available from their original sources. The LIDC-IDRI dataset is available from The Cancer Imaging Archive at \url{https://www.cancerimagingarchive.net/collection/lidc-idri/}. The RIGA dataset is available through Deep Blue Data at \url{https://deepblue.lib.umich.edu/data/concern/data_sets/3b591905z}. GleasonXAI image data are available from Harvard Dataverse at \url{https://dataverse.harvard.edu/dataset.xhtml?persistentId=doi:10.7910/DVN/OCYCMP}, and the corresponding multi-rater annotations are available from Figshare at \url{https://springernature.figshare.com/articles/dataset/Pathologist-like_explainable_AI_for_interpretable_Gleason_grading_in_prostate_cancer/27301845}. The MouseTumor dataset is described in Scientific Data and available through the resources listed in the publication at \url{https://www.nature.com/articles/s41597-024-03814-y#Tab3}. The MMIA dataset is available via Synapse at \url{https://www.synapse.org/Synapse:syn60868042/wiki/628716}. The MMIS dataset is available from the MMIS 2024 challenge website at \url{https://mmis2024.vercel.app/}.

\item \textbf{Code availability:} The benchmark suite code, including scripts for dataset integration, federated training, evaluation, and reproduction of the comparative benchmark, is publicly available at \url{https://github.com/MIC-DKFZ/FedSegNoiseBench}.

\end{itemize} 

\clearpage

\begin{appendices}

\section{Evaluation}\label{app0}

\subsection{Evaluation metric edge-case handling}
\label{app0:metric_edgecases}
Explicit edge-case handling is required in label-noisy segmentation benchmarks to avoid undefined metric behavior while ensuring that missing, spurious, or empty predictions are neither unfairly rewarded nor insufficiently penalized.

For \textit{Dice}, classes absent in both ground truth and prediction are assigned \texttt{NaN} and ignored during \texttt{nanmean}-based aggregation. If only one mask contains the class, or if both masks are non-empty but non-overlapping, Dice is set to 0.

For \textit{HD95}, empty masks are handled explicitly: if both masks are empty, HD95 is set to 0; if only one mask is empty, HD95 is set to 1000 as a fixed large penalty. Otherwise, HD95 is computed from symmetric surface distances using physical voxel spacing, with unit spacing used when spacing metadata is missing or inconsistent.

For \textit{foreground-background instance-level F1}, all non-background labels are merged into a foreground mask before instance extraction. If both masks contain no foreground instances, the score is \texttt{NaN}; if only one mask contains foreground instances, the score is 0. Instance matching is one-to-one with an IoU threshold of 0.1, thereby penalizing split or merged objects through unmatched components.

For \textit{class confusion}, the score is defined only for foreground classes present in the ground truth; absent classes are assigned \texttt{NaN}. Background predictions on foreground regions and foreground predictions in background regions are not counted as class confusion. Only foreground-to-foreground swaps are counted, so the metric is \texttt{NaN} for single-foreground-class tasks.

\clearpage

\section{Dataset- and method-specific training details}\label{app1}

\subsection{Dataset-specific nnU-Net hyperparameters}
\label{app1:data_hparams}

The benchmark suite's federated segmentation framework is based on the self-configuring nnU-Net framework, which derives experiment-planning and training hyperparameters from the processed dataset characteristics. Table~\ref{tab:appendix_nnunet_hparams} summarizes the dataset-specific batch size, patch size, median image size, and target spacing used in the experiments. Notably, we had to adjust the batch size to 1 for the \textit{FedSelect} experiments due to computational limitations.

\begin{table*}[h!]
\centering

\caption{Dataset-specific nnU-Net experiment-planning hyperparameters used in the federated benchmark, including batch size, patch size, median image size, and target spacing.}

\footnotesize
\setlength{\tabcolsep}{4pt}
\renewcommand{\arraystretch}{1.12}

\begin{adjustbox}{max width=\textwidth}
\begin{tabular}{
    L{2.8cm}
    *{6}{C{2.0cm}}
}
\toprule
& 
\textbf{LIDC} &
\textbf{RIGA} &
\textbf{GleasonHD} &
\textbf{MouseT} &
\textbf{MMIS} &
\textbf{MMIA} \\
\midrule

\textit{Batch size}
& 20
& 12
& 2 (1)
& 2
& 4
& 2 \\

\textit{Patch size}
& \makecell[c]{64, 64, 64}
& \makecell[c]{512, 512}
& \makecell[c]{2048, 2048}
& \makecell[c]{256, 96, 96}
& \makecell[c]{24, 192, 160}
& \makecell[c]{56, 192, 160}\\

\textit{Median image size (voxels)}
& \makecell[c]{64, 64, 64}
& \makecell[c]{1458, 1458}
& \makecell[c]{3100, 3100}
& \makecell[c]{480, 192, 192}
& \makecell[c]{24, 187, 147}
& \makecell[c]{80, 256, 256} \\

\textit{Target spacing}
& \makecell[c]{1.0, 1.0, 1.0}
& \makecell[c]{1.0, 1.0}
& \makecell[c]{1.0, 1.0}
& \makecell[c]{0.21, 0.21, 0.21}
& \makecell[c]{3.0, 0.508, 0.508}
& \makecell[c]{2.0, 0.703, 0.703}\\

\bottomrule
\end{tabular}
\end{adjustbox}

\label{tab:appendix_nnunet_hparams}
\end{table*}

\subsection{FNLL method-specific hyperparameters}
\label{app1:fnll_hparams}
For each compared FNLL method, hyperparameters were optimized by grid search over predefined candidate values. Final settings were selected according to the best average validation performance across all benchmark datasets, as summarized in Table~\ref{tab:fnll_hparams}.

\begin{table*}[h!]
\centering

\caption{Grid-searched FNLL hyperparameters. Final values in bold denote the final selected settings based on the best average validation performance across all benchmark datasets.}

\footnotesize
\setlength{\tabcolsep}{4pt}
\renewcommand{\arraystretch}{1.12}

\begin{tabular}{L{1.5cm} L{4.5cm} L{3.5cm}}
\toprule
\textbf{Method} & \textbf{Hyperparameter} & \textbf{Grid search values} \\
\midrule

\multirow{2}{*}{\textit{FedA3I}}
& \texttt{feda3i\_warmup\_rounds\_frac} &  \textbf{0.05}, 0.1, 0.2 \\
& \texttt{feda3i\_interw} &  \textbf{0.3}, 0.5, 0.7 \\
\midrule

\textit{IOP-FL}
& \texttt{iopfl\_alpha} & 0.05, 0.1,  \textbf{0.2}, 0.4, 0.6 \\
\midrule

\multirow{4}{*}{\textit{FedCorr}}
& \texttt{fedcorr\_preproc\_rounds\_frac} &  \textbf{0.05}, 0.1, 0.2 \\
& \texttt{fedcorr\_relabel\_ratio} &  \textbf{0.3}, 0.5, 0.7 \\
& \texttt{fedcorr\_relabel\_confidence\_thres} & 0.3,  \textbf{0.5}, 0.7 \\
& \texttt{fedcorr\_proxterm\_beta} & 3,  \textbf{5}, 8 \\
\midrule

\multirow{5}{*}{\textit{FedSelect}}
& \texttt{fedselect\_warmup\_rounds\_frac} & 0.05,  \textbf{0.1}, 0.2 \\
& \texttt{fedselect\_client\_select\_ratio} & 0.6, 0.8,  \textbf{1.0} \\
& \texttt{fedselect\_sample\_select\_ratio} & 0.5,  \textbf{0.75}, 1.0 \\
& \texttt{fedselect\_meta\_momentum} & 0.3, 0.5,  \textbf{0.7} \\
& \texttt{fedselect\_reward\_data\_size\_frac} & 0.05,  \textbf{0.1}, 0.2 \\
\bottomrule
\end{tabular}

\label{tab:fnll_hparams}
\end{table*}

\clearpage

\section{Detailed experimental results}\label{app2}

\subsection{General segmentation performance evaluation}
\label{app2:general_dice}

\begin{figure}[h!]
\centering
\includegraphics[width=\textwidth]{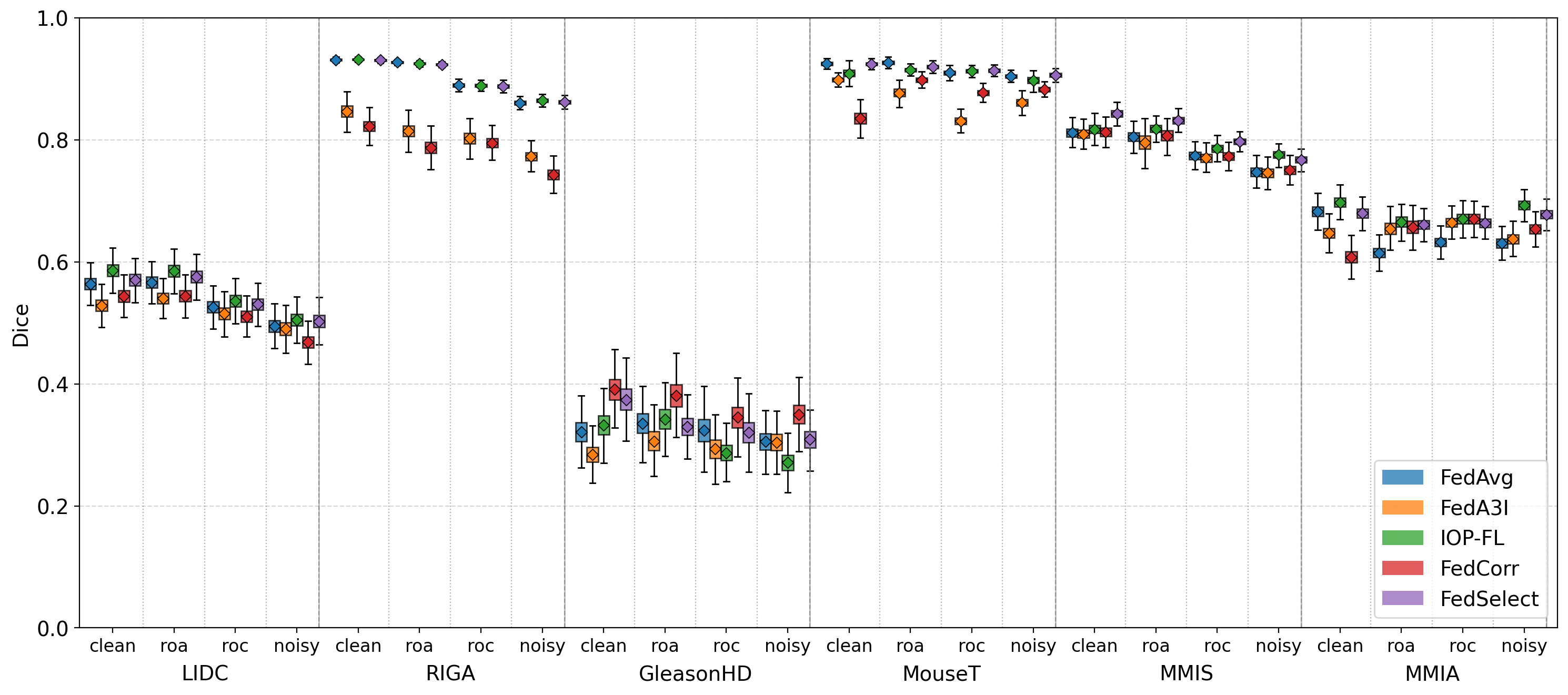}
\caption{Bootstrap Dice score distributions of all benchmarked methods across client-noise scenarios and all datasets. Black bars denote medians, and diamonds denote means.}
\label{fig:app_dice}
\end{figure}

General segmentation performance was assessed using the Dice score. Detailed results are reported in Table~\ref{tab:bootstrap_dice_results} and Figure~\ref{fig:app_dice}.

\begin{table*}[t]
\centering

\caption{Mean validation Dice values ($\times 100$) with 95\% percentile bootstrap confidence intervals for each dataset, noise scenario, and method.}

\small
\begin{tabular}{llccccc}
\hline
\hline
\textit{Dataset} & \textit{Scenario} & \textbf{FedAvg} & \textbf{FedA3I} & \textbf{IOP-FL} & \textbf{FedCorr} & \textbf{FedSelect} \\
\hline
\multirow{4}{*}{LIDC} & clean & \cellcolor[HTML]{E3F3B6}$56.4^{\scriptscriptstyle 59.0}_{\scriptscriptstyle 53.9}$ & \cellcolor[HTML]{C9667D}$52.9^{\scriptscriptstyle 55.5}_{\scriptscriptstyle 50.1}$ & \cellcolor[HTML]{66A487}$\mathbf{58.6}^{\scriptscriptstyle 61.5}_{\scriptscriptstyle 56.0}$ & \cellcolor[HTML]{FCC099}$54.4^{\scriptscriptstyle 57.1}_{\scriptscriptstyle 51.7}$ & \cellcolor[HTML]{BDE3A4}$\underline{57.1}^{\scriptscriptstyle 59.7}_{\scriptscriptstyle 54.5}$ \\
& roa & \cellcolor[HTML]{EDF7C0}$56.7^{\scriptscriptstyle 59.3}_{\scriptscriptstyle 54.1}$ & \cellcolor[HTML]{C9667D}$54.1^{\scriptscriptstyle 56.6}_{\scriptscriptstyle 51.6}$ & \cellcolor[HTML]{66A487}$\mathbf{58.5}^{\scriptscriptstyle 61.4}_{\scriptscriptstyle 55.8}$ & \cellcolor[HTML]{DE7A7D}$54.4^{\scriptscriptstyle 57.1}_{\scriptscriptstyle 51.8}$ & \cellcolor[HTML]{A9DAA2}$\underline{57.6}^{\scriptscriptstyle 60.5}_{\scriptscriptstyle 54.9}$ \\
& roc & \cellcolor[HTML]{E9F6BB}$52.6^{\scriptscriptstyle 55.2}_{\scriptscriptstyle 49.8}$ & \cellcolor[HTML]{F9A98F}$51.6^{\scriptscriptstyle 54.3}_{\scriptscriptstyle 48.9}$ & \cellcolor[HTML]{66A487}$\mathbf{53.6}^{\scriptscriptstyle 56.4}_{\scriptscriptstyle 50.7}$ & \cellcolor[HTML]{C9667D}$51.1^{\scriptscriptstyle 53.6}_{\scriptscriptstyle 48.4}$ & \cellcolor[HTML]{A1D7A1}$\underline{53.1}^{\scriptscriptstyle 55.8}_{\scriptscriptstyle 50.3}$ \\
& noisy & \cellcolor[HTML]{C4E6A5}$49.5^{\scriptscriptstyle 52.3}_{\scriptscriptstyle 46.7}$ & \cellcolor[HTML]{E9F6BB}$49.1^{\scriptscriptstyle 52.1}_{\scriptscriptstyle 46.1}$ & \cellcolor[HTML]{66A487}$\mathbf{50.5}^{\scriptscriptstyle 53.6}_{\scriptscriptstyle 47.7}$ & \cellcolor[HTML]{C9667D}$46.9^{\scriptscriptstyle 49.6}_{\scriptscriptstyle 44.3}$ & \cellcolor[HTML]{72BA92}$\underline{50.3}^{\scriptscriptstyle 53.1}_{\scriptscriptstyle 47.4}$ \\
\hline
\multirow{4}{*}{RIGA} & clean & \cellcolor[HTML]{67A688}$\underline{93.2}^{\scriptscriptstyle 93.5}_{\scriptscriptstyle 92.8}$ & \cellcolor[HTML]{FAB192}$84.7^{\scriptscriptstyle 87.1}_{\scriptscriptstyle 82.0}$ & \cellcolor[HTML]{66A487}$\mathbf{93.2}^{\scriptscriptstyle 93.6}_{\scriptscriptstyle 92.9}$ & \cellcolor[HTML]{C9667D}$82.2^{\scriptscriptstyle 84.4}_{\scriptscriptstyle 79.8}$ & \cellcolor[HTML]{68A889}$93.1^{\scriptscriptstyle 93.4}_{\scriptscriptstyle 92.7}$ \\
& roa & \cellcolor[HTML]{66A487}$\mathbf{92.8}^{\scriptscriptstyle 93.1}_{\scriptscriptstyle 92.4}$ & \cellcolor[HTML]{F8A68E}$81.5^{\scriptscriptstyle 83.8}_{\scriptscriptstyle 79.1}$ & \cellcolor[HTML]{68A889}$\underline{92.6}^{\scriptscriptstyle 93.0}_{\scriptscriptstyle 92.2}$ & \cellcolor[HTML]{C9667D}$78.8^{\scriptscriptstyle 81.3}_{\scriptscriptstyle 76.0}$ & \cellcolor[HTML]{6AAB8B}$92.4^{\scriptscriptstyle 92.8}_{\scriptscriptstyle 92.0}$ \\
& roc & \cellcolor[HTML]{66A487}$\mathbf{89.0}^{\scriptscriptstyle 89.7}_{\scriptscriptstyle 88.2}$ & \cellcolor[HTML]{DF7B7D}$80.3^{\scriptscriptstyle 82.7}_{\scriptscriptstyle 77.6}$ & \cellcolor[HTML]{67A688}$\underline{88.9}^{\scriptscriptstyle 89.6}_{\scriptscriptstyle 88.2}$ & \cellcolor[HTML]{C9667D}$79.5^{\scriptscriptstyle 81.6}_{\scriptscriptstyle 77.4}$ & \cellcolor[HTML]{68A989}$88.8^{\scriptscriptstyle 89.6}_{\scriptscriptstyle 88.1}$ \\
& noisy & \cellcolor[HTML]{6BAD8C}$86.1^{\scriptscriptstyle 86.9}_{\scriptscriptstyle 85.2}$ & \cellcolor[HTML]{FBB896}$77.3^{\scriptscriptstyle 79.2}_{\scriptscriptstyle 75.5}$ & \cellcolor[HTML]{66A487}$\mathbf{86.5}^{\scriptscriptstyle 87.2}_{\scriptscriptstyle 85.7}$ & \cellcolor[HTML]{C9667D}$74.4^{\scriptscriptstyle 76.7}_{\scriptscriptstyle 72.1}$ & \cellcolor[HTML]{69AA8A}$\underline{86.2}^{\scriptscriptstyle 87.0}_{\scriptscriptstyle 85.3}$ \\
\hline
\multirow{4}{*}{GleasonHD} & clean & \cellcolor[HTML]{FEDCAC}$32.1^{\scriptscriptstyle 36.4}_{\scriptscriptstyle 27.4}$ & \cellcolor[HTML]{C9667D}$28.4^{\scriptscriptstyle 31.7}_{\scriptscriptstyle 24.8}$ & \cellcolor[HTML]{FFF6C9}$33.2^{\scriptscriptstyle 37.2}_{\scriptscriptstyle 29.0}$ & \cellcolor[HTML]{66A487}$\mathbf{39.1}^{\scriptscriptstyle 43.8}_{\scriptscriptstyle 34.5}$ & \cellcolor[HTML]{8ECD9C}$\underline{37.4}^{\scriptscriptstyle 42.4}_{\scriptscriptstyle 32.7}$ \\
& roa & \cellcolor[HTML]{FEEAB7}$33.5^{\scriptscriptstyle 38.0}_{\scriptscriptstyle 29.0}$ & \cellcolor[HTML]{C9667D}$30.6^{\scriptscriptstyle 34.8}_{\scriptscriptstyle 26.3}$ & \cellcolor[HTML]{FFFCD3}$\underline{34.2}^{\scriptscriptstyle 38.7}_{\scriptscriptstyle 30.0}$ & \cellcolor[HTML]{66A487}$\mathbf{38.1}^{\scriptscriptstyle 43.1}_{\scriptscriptstyle 32.9}$ & \cellcolor[HTML]{FED5A6}$33.0^{\scriptscriptstyle 36.9}_{\scriptscriptstyle 29.2}$ \\
& roc & \cellcolor[HTML]{DDF1B2}$\underline{32.4}^{\scriptscriptstyle 37.4}_{\scriptscriptstyle 27.5}$ & \cellcolor[HTML]{E9867F}$29.4^{\scriptscriptstyle 33.2}_{\scriptscriptstyle 25.5}$ & \cellcolor[HTML]{C9667D}$28.7^{\scriptscriptstyle 32.4}_{\scriptscriptstyle 25.0}$ & \cellcolor[HTML]{66A487}$\mathbf{34.5}^{\scriptscriptstyle 39.2}_{\scriptscriptstyle 29.9}$ & \cellcolor[HTML]{EDF7C0}$32.1^{\scriptscriptstyle 36.8}_{\scriptscriptstyle 27.3}$ \\
& noisy & \cellcolor[HTML]{FFF4C6}$30.6^{\scriptscriptstyle 34.4}_{\scriptscriptstyle 26.7}$ & \cellcolor[HTML]{FFF0C0}$30.4^{\scriptscriptstyle 34.2}_{\scriptscriptstyle 26.5}$ & \cellcolor[HTML]{C9667D}$27.1^{\scriptscriptstyle 30.6}_{\scriptscriptstyle 23.5}$ & \cellcolor[HTML]{66A487}$\mathbf{35.0}^{\scriptscriptstyle 39.2}_{\scriptscriptstyle 30.7}$ & \cellcolor[HTML]{FFFBD2}$\underline{30.9}^{\scriptscriptstyle 34.4}_{\scriptscriptstyle 27.0}$ \\
\hline
\multirow{4}{*}{MouseT} & clean & \cellcolor[HTML]{66A487}$\mathbf{92.5}^{\scriptscriptstyle 93.2}_{\scriptscriptstyle 91.9}$ & \cellcolor[HTML]{C9E8A6}$89.9^{\scriptscriptstyle 90.7}_{\scriptscriptstyle 88.9}$ & \cellcolor[HTML]{98D29F}$90.9^{\scriptscriptstyle 92.2}_{\scriptscriptstyle 89.2}$ & \cellcolor[HTML]{C9667D}$83.6^{\scriptscriptstyle 85.7}_{\scriptscriptstyle 81.1}$ & \cellcolor[HTML]{67A688}$\underline{92.5}^{\scriptscriptstyle 93.1}_{\scriptscriptstyle 91.8}$ \\
& roa & \cellcolor[HTML]{66A487}$\mathbf{92.7}^{\scriptscriptstyle 93.3}_{\scriptscriptstyle 92.0}$ & \cellcolor[HTML]{C9667D}$87.7^{\scriptscriptstyle 89.4}_{\scriptscriptstyle 85.9}$ & \cellcolor[HTML]{AFDDA3}$91.5^{\scriptscriptstyle 92.3}_{\scriptscriptstyle 90.8}$ & \cellcolor[HTML]{FFF3C4}$89.9^{\scriptscriptstyle 90.8}_{\scriptscriptstyle 88.8}$ & \cellcolor[HTML]{87C99A}$\underline{92.0}^{\scriptscriptstyle 92.7}_{\scriptscriptstyle 91.3}$ \\
& roc & \cellcolor[HTML]{6DB18D}$91.0^{\scriptscriptstyle 91.9}_{\scriptscriptstyle 90.1}$ & \cellcolor[HTML]{C9667D}$83.1^{\scriptscriptstyle 84.5}_{\scriptscriptstyle 81.6}$ & \cellcolor[HTML]{68A889}$\underline{91.3}^{\scriptscriptstyle 92.0}_{\scriptscriptstyle 90.6}$ & \cellcolor[HTML]{F1F9C6}$87.8^{\scriptscriptstyle 88.9}_{\scriptscriptstyle 86.5}$ & \cellcolor[HTML]{66A487}$\mathbf{91.4}^{\scriptscriptstyle 92.2}_{\scriptscriptstyle 90.6}$ \\
& noisy & \cellcolor[HTML]{6CAF8C}$\underline{90.5}^{\scriptscriptstyle 91.3}_{\scriptscriptstyle 89.7}$ & \cellcolor[HTML]{C9667D}$86.1^{\scriptscriptstyle 87.5}_{\scriptscriptstyle 84.7}$ & \cellcolor[HTML]{9ED5A0}$89.8^{\scriptscriptstyle 91.0}_{\scriptscriptstyle 88.3}$ & \cellcolor[HTML]{FFFBD2}$88.3^{\scriptscriptstyle 89.2}_{\scriptscriptstyle 87.3}$ & \cellcolor[HTML]{66A487}$\mathbf{90.7}^{\scriptscriptstyle 91.5}_{\scriptscriptstyle 89.8}$ \\
\hline
\multirow{4}{*}{MMIS} & clean & \cellcolor[HTML]{DB777D}$81.2^{\scriptscriptstyle 83.0}_{\scriptscriptstyle 79.3}$ & \cellcolor[HTML]{C9667D}$81.0^{\scriptscriptstyle 82.8}_{\scriptscriptstyle 79.1}$ & \cellcolor[HTML]{FAB494}$\underline{81.8}^{\scriptscriptstyle 83.7}_{\scriptscriptstyle 79.7}$ & \cellcolor[HTML]{E37F7D}$81.3^{\scriptscriptstyle 83.1}_{\scriptscriptstyle 79.3}$ & \cellcolor[HTML]{66A487}$\mathbf{84.3}^{\scriptscriptstyle 85.6}_{\scriptscriptstyle 82.9}$ \\
& roa & \cellcolor[HTML]{FBBB97}$80.5^{\scriptscriptstyle 82.3}_{\scriptscriptstyle 78.7}$ & \cellcolor[HTML]{C9667D}$79.6^{\scriptscriptstyle 82.5}_{\scriptscriptstyle 76.6}$ & \cellcolor[HTML]{E1F2B5}$\underline{81.9}^{\scriptscriptstyle 83.5}_{\scriptscriptstyle 80.2}$ & \cellcolor[HTML]{FED1A3}$80.7^{\scriptscriptstyle 82.9}_{\scriptscriptstyle 78.4}$ & \cellcolor[HTML]{66A487}$\mathbf{83.2}^{\scriptscriptstyle 84.6}_{\scriptscriptstyle 81.7}$ \\
& roc & \cellcolor[HTML]{EB8C82}$77.4^{\scriptscriptstyle 79.2}_{\scriptscriptstyle 75.7}$ & \cellcolor[HTML]{C9667D}$77.1^{\scriptscriptstyle 78.9}_{\scriptscriptstyle 75.3}$ & \cellcolor[HTML]{F0F9C4}$\underline{78.6}^{\scriptscriptstyle 80.1}_{\scriptscriptstyle 76.9}$ & \cellcolor[HTML]{E6827D}$77.3^{\scriptscriptstyle 79.1}_{\scriptscriptstyle 75.6}$ & \cellcolor[HTML]{66A487}$\mathbf{79.8}^{\scriptscriptstyle 81.0}_{\scriptscriptstyle 78.5}$ \\
& noisy & \cellcolor[HTML]{D6727D}$74.8^{\scriptscriptstyle 76.8}_{\scriptscriptstyle 72.9}$ & \cellcolor[HTML]{C9667D}$74.6^{\scriptscriptstyle 76.7}_{\scriptscriptstyle 72.8}$ & \cellcolor[HTML]{66A487}$\mathbf{77.6}^{\scriptscriptstyle 79.0}_{\scriptscriptstyle 75.9}$ & \cellcolor[HTML]{EF9385}$75.1^{\scriptscriptstyle 77.0}_{\scriptscriptstyle 73.2}$ & \cellcolor[HTML]{C1E5A5}$\underline{76.7}^{\scriptscriptstyle 78.0}_{\scriptscriptstyle 75.3}$ \\
\hline
\multirow{4}{*}{MMIA} & clean & \cellcolor[HTML]{93D09D}$\underline{68.3}^{\scriptscriptstyle 70.4}_{\scriptscriptstyle 66.0}$ & \cellcolor[HTML]{FFF4C7}$64.8^{\scriptscriptstyle 67.0}_{\scriptscriptstyle 62.5}$ & \cellcolor[HTML]{66A487}$\mathbf{69.8}^{\scriptscriptstyle 71.8}_{\scriptscriptstyle 67.7}$ & \cellcolor[HTML]{C9667D}$60.8^{\scriptscriptstyle 63.5}_{\scriptscriptstyle 58.1}$ & \cellcolor[HTML]{A1D7A1}$68.0^{\scriptscriptstyle 70.1}_{\scriptscriptstyle 66.0}$ \\
& roa & \cellcolor[HTML]{C9667D}$61.5^{\scriptscriptstyle 63.8}_{\scriptscriptstyle 59.1}$ & \cellcolor[HTML]{A9DAA2}$65.5^{\scriptscriptstyle 68.2}_{\scriptscriptstyle 62.8}$ & \cellcolor[HTML]{66A487}$\mathbf{66.6}^{\scriptscriptstyle 68.8}_{\scriptscriptstyle 64.3}$ & \cellcolor[HTML]{97D19E}$65.7^{\scriptscriptstyle 68.4}_{\scriptscriptstyle 62.6}$ & \cellcolor[HTML]{74BE95}$\underline{66.1}^{\scriptscriptstyle 68.2}_{\scriptscriptstyle 64.1}$ \\
& roc & \cellcolor[HTML]{C9667D}$63.3^{\scriptscriptstyle 65.1}_{\scriptscriptstyle 61.3}$ & \cellcolor[HTML]{91CF9D}$66.4^{\scriptscriptstyle 68.5}_{\scriptscriptstyle 64.2}$ & \cellcolor[HTML]{66A487}$\mathbf{67.1}^{\scriptscriptstyle 69.3}_{\scriptscriptstyle 64.8}$ & \cellcolor[HTML]{66A487}$\underline{67.1}^{\scriptscriptstyle 69.3}_{\scriptscriptstyle 64.7}$ & \cellcolor[HTML]{95D09E}$66.4^{\scriptscriptstyle 68.4}_{\scriptscriptstyle 64.4}$ \\
& noisy & \cellcolor[HTML]{C9667D}$63.1^{\scriptscriptstyle 65.4}_{\scriptscriptstyle 61.0}$ & \cellcolor[HTML]{E7847E}$63.8^{\scriptscriptstyle 66.0}_{\scriptscriptstyle 61.6}$ & \cellcolor[HTML]{66A487}$\mathbf{69.3}^{\scriptscriptstyle 71.1}_{\scriptscriptstyle 67.3}$ & \cellcolor[HTML]{FEE4B2}$65.4^{\scriptscriptstyle 67.8}_{\scriptscriptstyle 63.1}$ & \cellcolor[HTML]{B4DFA3}$\underline{67.8}^{\scriptscriptstyle 69.7}_{\scriptscriptstyle 65.8}$ \\
\hline
\hline
\end{tabular}

\label{tab:bootstrap_dice_results_w_ci}
\end{table*}

\clearpage

\subsection{Noise-specific segmentation performance evaluation}
\label{app2:noisespecific}

\subsubsection{Contour-based segmentation label noise}
\label{app2:noisespecific_hd95}

\begin{figure}[h!]
\centering
\includegraphics[width=\textwidth]{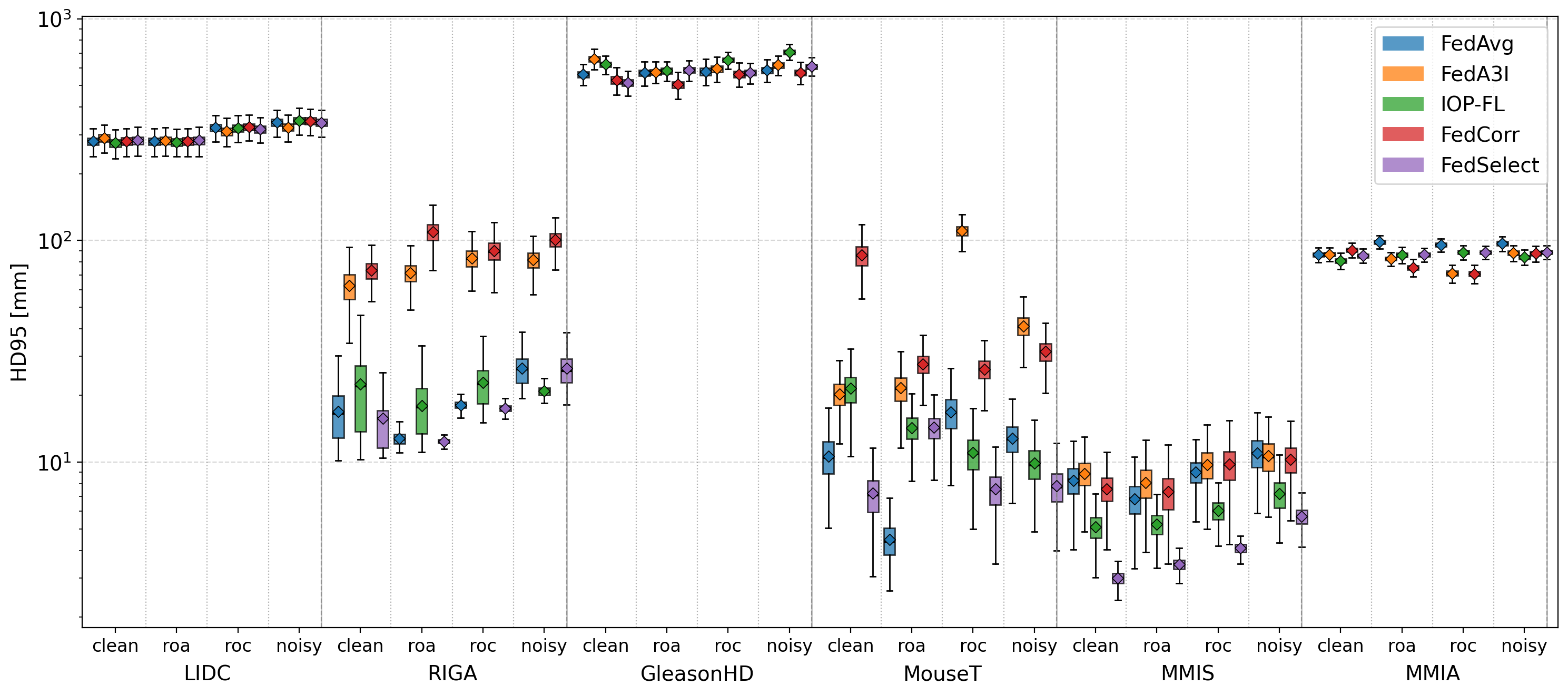}
\caption{Bootstrap HD95 distributions of all benchmarked methods across client-noise scenarios and all datasets. Black bars denote medians, and diamonds denote means.}
\label{fig:app_hd95}
\end{figure}

\begin{figure}[h!]
\centering
\includegraphics[width=\textwidth]{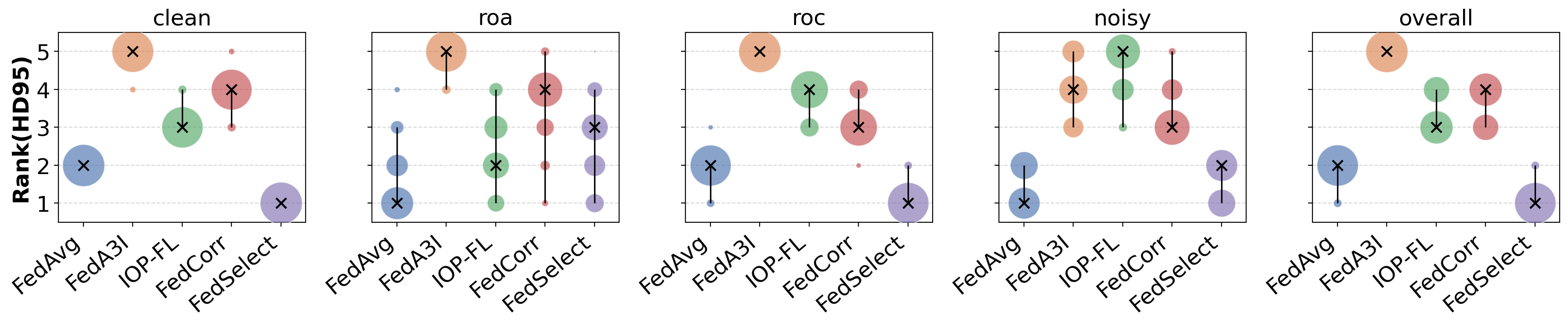}
\caption{Rank stability of HD95 segmentation performances based on bootstrapping across client-noise scenarios and methods on LIDC, RIDA, GleasonHD, MouseT, MMIS, and MMIA. Lower ranks indicate better performances, bubble size reflects the ranking frequency across datasets.}
\label{fig:app_ranking_hd95}
\end{figure}

Contour-related segmentation label noise was assessed using the HD95 metric on datasets featuring this noise type. Detailed results are reported in Table~\ref{tab:bootstrap_hd95_results} and Figures~\ref{fig:app_hd95},~\ref{fig:app_ranking_hd95}.

\begin{table*}[h!]
\centering

\caption{Mean validation HD95 values with 95\% percentile bootstrap confidence intervals for each dataset, noise scenario, and method.}

\small
\begin{tabular}{llccccc}
\hline
\hline
\textit{Dataset} & \textit{Scenario} & \textbf{FedAvg} & \textbf{FedA3I} & \textbf{IOP-FL} & \textbf{FedCorr} & \textbf{FedSelect} \\
\hline
\multirow{4}{*}{LIDC} & clean & \cellcolor[HTML]{CDE9A8}$\underline{279.9}^{\scriptscriptstyle 312.6}_{\scriptscriptstyle 249.3}$ & \cellcolor[HTML]{C9667D}$289.8^{\scriptscriptstyle 321.2}_{\scriptscriptstyle 260.1}$ & \cellcolor[HTML]{66A487}$\mathbf{275.4}^{\scriptscriptstyle 308.7}_{\scriptscriptstyle 244.4}$ & \cellcolor[HTML]{DCF0B2}$280.6^{\scriptscriptstyle 313.2}_{\scriptscriptstyle 250.1}$ & \cellcolor[HTML]{FFFDD6}$282.7^{\scriptscriptstyle 315.4}_{\scriptscriptstyle 252.7}$ \\
& roa & \cellcolor[HTML]{F9FDD1}$\underline{280.2}^{\scriptscriptstyle 312.8}_{\scriptscriptstyle 249.8}$ & \cellcolor[HTML]{F19887}$282.0^{\scriptscriptstyle 314.3}_{\scriptscriptstyle 251.7}$ & \cellcolor[HTML]{66A487}$\mathbf{278.0}^{\scriptscriptstyle 310.7}_{\scriptscriptstyle 247.3}$ & \cellcolor[HTML]{FFFBD2}$280.4^{\scriptscriptstyle 313.0}_{\scriptscriptstyle 250.1}$ & \cellcolor[HTML]{C9667D}$282.7^{\scriptscriptstyle 315.4}_{\scriptscriptstyle 252.7}$ \\
& roc & \cellcolor[HTML]{F09586}$322.7^{\scriptscriptstyle 355.9}_{\scriptscriptstyle 291.5}$ & \cellcolor[HTML]{66A487}$\mathbf{310.4}^{\scriptscriptstyle 343.6}_{\scriptscriptstyle 278.6}$ & \cellcolor[HTML]{FBBB97}$321.2^{\scriptscriptstyle 354.4}_{\scriptscriptstyle 289.6}$ & \cellcolor[HTML]{C9667D}$324.9^{\scriptscriptstyle 357.9}_{\scriptscriptstyle 292.9}$ & \cellcolor[HTML]{FDFED6}$\underline{317.5}^{\scriptscriptstyle 350.3}_{\scriptscriptstyle 285.4}$ \\
& noisy & \cellcolor[HTML]{FBBA97}$341.4^{\scriptscriptstyle 375.9}_{\scriptscriptstyle 307.0}$ & \cellcolor[HTML]{66A487}$\mathbf{324.1}^{\scriptscriptstyle 357.8}_{\scriptscriptstyle 289.2}$ & \cellcolor[HTML]{C9667D}$347.1^{\scriptscriptstyle 380.6}_{\scriptscriptstyle 309.8}$ & \cellcolor[HTML]{DD797D}$345.5^{\scriptscriptstyle 379.3}_{\scriptscriptstyle 308.4}$ & \cellcolor[HTML]{FED1A3}$\underline{340.0}^{\scriptscriptstyle 374.5}_{\scriptscriptstyle 306.1}$ \\
\hline
\multirow{4}{*}{RIGA} & clean & \cellcolor[HTML]{69AA8A}$\underline{16.9}^{\scriptscriptstyle 27.2}_{\scriptscriptstyle 11.1}$ & \cellcolor[HTML]{F6A28C}$62.6^{\scriptscriptstyle 85.8}_{\scriptscriptstyle 41.3}$ & \cellcolor[HTML]{7CC498}$22.4^{\scriptscriptstyle 43.4}_{\scriptscriptstyle 11.4}$ & \cellcolor[HTML]{C9667D}$73.4^{\scriptscriptstyle 91.2}_{\scriptscriptstyle 57.2}$ & \cellcolor[HTML]{66A487}$\mathbf{15.8}^{\scriptscriptstyle 25.3}_{\scriptscriptstyle 10.8}$ \\
& roa & \cellcolor[HTML]{67A688}$\underline{12.8}^{\scriptscriptstyle 15.1}_{\scriptscriptstyle 11.4}$ & \cellcolor[HTML]{FEEAB7}$71.5^{\scriptscriptstyle 88.7}_{\scriptscriptstyle 55.4}$ & \cellcolor[HTML]{6FB48F}$18.0^{\scriptscriptstyle 29.8}_{\scriptscriptstyle 11.8}$ & \cellcolor[HTML]{C9667D}$109.6^{\scriptscriptstyle 140.6}_{\scriptscriptstyle 82.7}$ & \cellcolor[HTML]{66A487}$\mathbf{12.4}^{\scriptscriptstyle 13.0}_{\scriptscriptstyle 11.7}$ \\
& roc & \cellcolor[HTML]{67A788}$\underline{18.1}^{\scriptscriptstyle 19.8}_{\scriptscriptstyle 16.5}$ & \cellcolor[HTML]{E5817D}$83.2^{\scriptscriptstyle 102.7}_{\scriptscriptstyle 64.3}$ & \cellcolor[HTML]{71B992}$22.9^{\scriptscriptstyle 34.2}_{\scriptscriptstyle 16.4}$ & \cellcolor[HTML]{C9667D}$90.0^{\scriptscriptstyle 116.0}_{\scriptscriptstyle 68.7}$ & \cellcolor[HTML]{66A487}$\mathbf{17.5}^{\scriptscriptstyle 19.0}_{\scriptscriptstyle 16.1}$ \\
& noisy & \cellcolor[HTML]{71B992}$26.5^{\scriptscriptstyle 37.3}_{\scriptscriptstyle 20.5}$ & \cellcolor[HTML]{FAB595}$81.9^{\scriptscriptstyle 101.3}_{\scriptscriptstyle 65.2}$ & \cellcolor[HTML]{66A487}$\mathbf{20.9}^{\scriptscriptstyle 23.3}_{\scriptscriptstyle 19.0}$ & \cellcolor[HTML]{C9667D}$100.9^{\scriptscriptstyle 121.3}_{\scriptscriptstyle 82.2}$ & \cellcolor[HTML]{70B891}$\underline{26.5}^{\scriptscriptstyle 38.3}_{\scriptscriptstyle 19.3}$ \\
\hline
\multirow{4}{*}{GleasonHD} & clean & \cellcolor[HTML]{D1ECAB}$561.7^{\scriptscriptstyle 613.4}_{\scriptscriptstyle 514.6}$ & \cellcolor[HTML]{C9667D}$659.1^{\scriptscriptstyle 711.0}_{\scriptscriptstyle 610.0}$ & \cellcolor[HTML]{FCBE99}$621.8^{\scriptscriptstyle 665.9}_{\scriptscriptstyle 579.6}$ & \cellcolor[HTML]{76C296}$\underline{529.6}^{\scriptscriptstyle 584.2}_{\scriptscriptstyle 477.6}$ & \cellcolor[HTML]{66A487}$\mathbf{514.9}^{\scriptscriptstyle 564.3}_{\scriptscriptstyle 466.8}$ \\
& roa & \cellcolor[HTML]{F8A68E}$\underline{570.0}^{\scriptscriptstyle 619.3}_{\scriptscriptstyle 522.4}$ & \cellcolor[HTML]{ED8F83}$575.2^{\scriptscriptstyle 626.7}_{\scriptscriptstyle 526.0}$ & \cellcolor[HTML]{D7747D}$582.1^{\scriptscriptstyle 627.1}_{\scriptscriptstyle 539.9}$ & \cellcolor[HTML]{66A487}$\mathbf{505.0}^{\scriptscriptstyle 555.6}_{\scriptscriptstyle 457.9}$ & \cellcolor[HTML]{C9667D}$586.0^{\scriptscriptstyle 629.5}_{\scriptscriptstyle 540.6}$ \\
& roc & \cellcolor[HTML]{9AD39F}$577.9^{\scriptscriptstyle 636.3}_{\scriptscriptstyle 523.8}$ & \cellcolor[HTML]{E0F2B4}$595.0^{\scriptscriptstyle 651.6}_{\scriptscriptstyle 540.4}$ & \cellcolor[HTML]{C9667D}$650.9^{\scriptscriptstyle 695.7}_{\scriptscriptstyle 607.9}$ & \cellcolor[HTML]{66A487}$\mathbf{561.8}^{\scriptscriptstyle 611.7}_{\scriptscriptstyle 515.9}$ & \cellcolor[HTML]{7AC397}$\underline{571.6}^{\scriptscriptstyle 620.4}_{\scriptscriptstyle 527.9}$ \\
& noisy & \cellcolor[HTML]{7AC397}$\underline{586.9}^{\scriptscriptstyle 636.6}_{\scriptscriptstyle 537.9}$ & \cellcolor[HTML]{D6EEAE}$618.4^{\scriptscriptstyle 665.5}_{\scriptscriptstyle 574.3}$ & \cellcolor[HTML]{C9667D}$708.4^{\scriptscriptstyle 755.8}_{\scriptscriptstyle 664.4}$ & \cellcolor[HTML]{66A487}$\mathbf{571.8}^{\scriptscriptstyle 620.5}_{\scriptscriptstyle 527.8}$ & \cellcolor[HTML]{C1E5A5}$610.0^{\scriptscriptstyle 655.2}_{\scriptscriptstyle 567.3}$ \\
\hline
\multirow{4}{*}{MouseT} & clean & \cellcolor[HTML]{6DB18D}$\underline{10.7}^{\scriptscriptstyle 15.5}_{\scriptscriptstyle 6.4}$ & \cellcolor[HTML]{93D09D}$20.3^{\scriptscriptstyle 26.7}_{\scriptscriptstyle 14.3}$ & \cellcolor[HTML]{9AD39F}$21.5^{\scriptscriptstyle 29.8}_{\scriptscriptstyle 13.6}$ & \cellcolor[HTML]{C9667D}$85.9^{\scriptscriptstyle 109.9}_{\scriptscriptstyle 65.5}$ & \cellcolor[HTML]{66A487}$\mathbf{7.3}^{\scriptscriptstyle 11.0}_{\scriptscriptstyle 4.4}$ \\
& roa & \cellcolor[HTML]{66A487}$\mathbf{4.5}^{\scriptscriptstyle 6.3}_{\scriptscriptstyle 3.1}$ & \cellcolor[HTML]{FCBE99}$21.6^{\scriptscriptstyle 29.2}_{\scriptscriptstyle 15.3}$ & \cellcolor[HTML]{EEF8C1}$\underline{14.3}^{\scriptscriptstyle 19.3}_{\scriptscriptstyle 9.8}$ & \cellcolor[HTML]{C9667D}$27.7^{\scriptscriptstyle 35.6}_{\scriptscriptstyle 20.1}$ & \cellcolor[HTML]{EEF8C1}$14.3^{\scriptscriptstyle 19.4}_{\scriptscriptstyle 10.0}$ \\
& roc & \cellcolor[HTML]{74BE95}$16.8^{\scriptscriptstyle 24.6}_{\scriptscriptstyle 10.4}$ & \cellcolor[HTML]{C9667D}$110.4^{\scriptscriptstyle 127.5}_{\scriptscriptstyle 94.5}$ & \cellcolor[HTML]{6BAD8C}$\underline{11.1}^{\scriptscriptstyle 16.3}_{\scriptscriptstyle 6.7}$ & \cellcolor[HTML]{9AD39F}$26.2^{\scriptscriptstyle 33.0}_{\scriptscriptstyle 18.8}$ & \cellcolor[HTML]{66A487}$\mathbf{7.5}^{\scriptscriptstyle 11.2}_{\scriptscriptstyle 4.7}$ \\
& noisy & \cellcolor[HTML]{8CCC9C}$12.8^{\scriptscriptstyle 18.3}_{\scriptscriptstyle 8.0}$ & \cellcolor[HTML]{C9667D}$41.2^{\scriptscriptstyle 52.5}_{\scriptscriptstyle 31.1}$ & \cellcolor[HTML]{70B690}$\underline{9.9}^{\scriptscriptstyle 14.6}_{\scriptscriptstyle 6.1}$ & \cellcolor[HTML]{FDC99E}$31.6^{\scriptscriptstyle 39.9}_{\scriptscriptstyle 23.5}$ & \cellcolor[HTML]{66A487}$\mathbf{7.8}^{\scriptscriptstyle 11.4}_{\scriptscriptstyle 4.8}$ \\
\hline
\multirow{4}{*}{MMIS} & clean & \cellcolor[HTML]{E7847E}$8.3^{\scriptscriptstyle 11.0}_{\scriptscriptstyle 5.3}$ & \cellcolor[HTML]{C9667D}$8.9^{\scriptscriptstyle 11.9}_{\scriptscriptstyle 6.0}$ & \cellcolor[HTML]{DBF0B1}$\underline{5.1}^{\scriptscriptstyle 6.7}_{\scriptscriptstyle 3.6}$ & \cellcolor[HTML]{FAB192}$7.6^{\scriptscriptstyle 9.9}_{\scriptscriptstyle 5.1}$ & \cellcolor[HTML]{66A487}$\mathbf{3.0}^{\scriptscriptstyle 3.4}_{\scriptscriptstyle 2.6}$ \\
& roa & \cellcolor[HTML]{FCC39B}$6.8^{\scriptscriptstyle 9.2}_{\scriptscriptstyle 4.5}$ & \cellcolor[HTML]{C9667D}$8.1^{\scriptscriptstyle 11.3}_{\scriptscriptstyle 5.2}$ & \cellcolor[HTML]{E5F4B7}$\underline{5.3}^{\scriptscriptstyle 6.7}_{\scriptscriptstyle 3.9}$ & \cellcolor[HTML]{F19887}$7.4^{\scriptscriptstyle 10.9}_{\scriptscriptstyle 4.4}$ & \cellcolor[HTML]{66A487}$\mathbf{3.5}^{\scriptscriptstyle 3.9}_{\scriptscriptstyle 3.0}$ \\
& roc & \cellcolor[HTML]{ED9084}$9.0^{\scriptscriptstyle 11.6}_{\scriptscriptstyle 6.4}$ & \cellcolor[HTML]{CB687D}$9.8^{\scriptscriptstyle 13.2}_{\scriptscriptstyle 6.6}$ & \cellcolor[HTML]{D6EEAE}$\underline{6.0}^{\scriptscriptstyle 7.5}_{\scriptscriptstyle 4.8}$ & \cellcolor[HTML]{C9667D}$9.8^{\scriptscriptstyle 14.6}_{\scriptscriptstyle 6.2}$ & \cellcolor[HTML]{66A487}$\mathbf{4.1}^{\scriptscriptstyle 4.5}_{\scriptscriptstyle 3.6}$ \\
& noisy & \cellcolor[HTML]{C9667D}$11.0^{\scriptscriptstyle 15.0}_{\scriptscriptstyle 7.2}$ & \cellcolor[HTML]{DB777D}$10.7^{\scriptscriptstyle 15.0}_{\scriptscriptstyle 7.1}$ & \cellcolor[HTML]{C3E5A5}$\underline{7.2}^{\scriptscriptstyle 9.8}_{\scriptscriptstyle 4.9}$ & \cellcolor[HTML]{EB8C82}$10.3^{\scriptscriptstyle 14.5}_{\scriptscriptstyle 6.9}$ & \cellcolor[HTML]{66A487}$\mathbf{5.7}^{\scriptscriptstyle 6.9}_{\scriptscriptstyle 4.6}$ \\
\hline
\multirow{4}{*}{MMIA} & clean & \cellcolor[HTML]{FFF5C8}$86.3^{\scriptscriptstyle 91.4}_{\scriptscriptstyle 81.4}$ & \cellcolor[HTML]{FEEFBE}$86.6^{\scriptscriptstyle 91.0}_{\scriptscriptstyle 81.9}$ & \cellcolor[HTML]{66A487}$\mathbf{81.1}^{\scriptscriptstyle 86.2}_{\scriptscriptstyle 75.9}$ & \cellcolor[HTML]{C9667D}$90.5^{\scriptscriptstyle 95.5}_{\scriptscriptstyle 85.1}$ & \cellcolor[HTML]{F3FAC8}$\underline{85.3}^{\scriptscriptstyle 90.3}_{\scriptscriptstyle 80.7}$ \\
& roa & \cellcolor[HTML]{C9667D}$98.4^{\scriptscriptstyle 104.1}_{\scriptscriptstyle 93.2}$ & \cellcolor[HTML]{CDE9A8}$\underline{82.6}^{\scriptscriptstyle 87.1}_{\scriptscriptstyle 77.9}$ & \cellcolor[HTML]{F6FBCC}$86.0^{\scriptscriptstyle 91.3}_{\scriptscriptstyle 80.2}$ & \cellcolor[HTML]{66A487}$\mathbf{75.5}^{\scriptscriptstyle 80.3}_{\scriptscriptstyle 70.4}$ & \cellcolor[HTML]{F8FCCF}$86.3^{\scriptscriptstyle 90.8}_{\scriptscriptstyle 81.8}$ \\
& roc & \cellcolor[HTML]{C9667D}$95.6^{\scriptscriptstyle 100.6}_{\scriptscriptstyle 90.4}$ & \cellcolor[HTML]{68A889}$\underline{71.1}^{\scriptscriptstyle 76.1}_{\scriptscriptstyle 65.9}$ & \cellcolor[HTML]{FDC99E}$88.5^{\scriptscriptstyle 93.4}_{\scriptscriptstyle 83.1}$ & \cellcolor[HTML]{66A487}$\mathbf{70.7}^{\scriptscriptstyle 75.8}_{\scriptscriptstyle 65.4}$ & \cellcolor[HTML]{FDCB9E}$88.3^{\scriptscriptstyle 93.2}_{\scriptscriptstyle 83.6}$ \\
& noisy & \cellcolor[HTML]{C9667D}$96.9^{\scriptscriptstyle 102.9}_{\scriptscriptstyle 91.4}$ & \cellcolor[HTML]{CAE8A6}$87.9^{\scriptscriptstyle 92.9}_{\scriptscriptstyle 82.5}$ & \cellcolor[HTML]{66A487}$\mathbf{83.9}^{\scriptscriptstyle 88.7}_{\scriptscriptstyle 78.9}$ & \cellcolor[HTML]{BDE3A4}$\underline{87.4}^{\scriptscriptstyle 92.4}_{\scriptscriptstyle 82.0}$ & \cellcolor[HTML]{D9EFAF}$88.5^{\scriptscriptstyle 93.1}_{\scriptscriptstyle 83.9}$ \\
\hline
\hline
\end{tabular}

\label{tab:bootstrap_hd95_results}
\end{table*}

\clearpage

\subsubsection{Instance-based segmentation label noise}
\label{app2:noisespecific_f1}

\begin{figure}[h!]
\centering
\includegraphics[width=\textwidth]{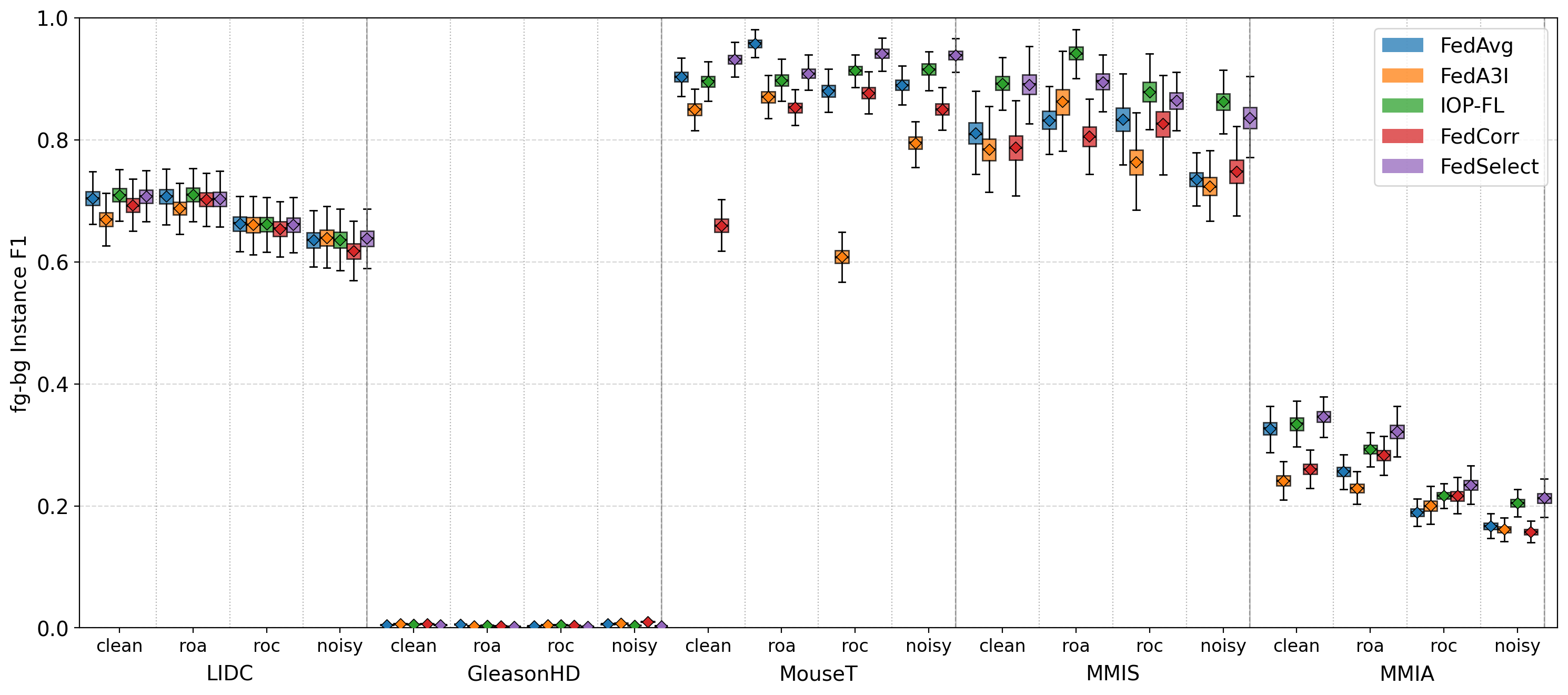}
\caption{Bootstrap foreground-background instance-level F1 score distributions of all benchmarked methods across client-noise scenarios and all datasets featuring instance-based noise (LIDC, GleasonHD, MouseT, MMIS, MMIA). Black bars denote medians, and diamonds denote means.}
\label{fig:app_fgbginstancef1}
\end{figure}

Instance-related segmentation label noise was assessed using the foreground-background instance-level F1 score on datasets featuring this noise type. Detailed results are reported in Table~\ref{tab:bootstrap_fgbginstancef1_results} and Figures~\ref{fig:app_fgbginstancef1},~\ref{fig:app_fgbginstancef1}.

\begin{table*}[h!]
\centering

\caption{Mean validation foreground-background instance-level F1 values ($\times 100$) with 95\% percentile bootstrap confidence intervals for each dataset, noise scenario, and method.}

\small
\begin{tabular}{llccccc}
\hline
\hline
\textit{Dataset} & \textit{Scenario} & \textbf{FedAvg} & \textbf{FedA3I} & \textbf{IOP-FL} & \textbf{FedCorr} & \textbf{FedSelect} \\
\hline
\multirow{4}{*}{LIDC} & clean & \cellcolor[HTML]{85C99A}$70.4^{\scriptscriptstyle 73.8}_{\scriptscriptstyle 67.2}$ & \cellcolor[HTML]{C9667D}$67.0^{\scriptscriptstyle 70.2}_{\scriptscriptstyle 63.4}$ & \cellcolor[HTML]{66A487}$\mathbf{71.0}^{\scriptscriptstyle 74.3}_{\scriptscriptstyle 67.8}$ & \cellcolor[HTML]{EEF8C1}$69.3^{\scriptscriptstyle 72.7}_{\scriptscriptstyle 66.0}$ & \cellcolor[HTML]{6FB48F}$\underline{70.7}^{\scriptscriptstyle 73.9}_{\scriptscriptstyle 67.5}$ \\
& roa & \cellcolor[HTML]{81C799}$\underline{70.8}^{\scriptscriptstyle 74.0}_{\scriptscriptstyle 67.5}$ & \cellcolor[HTML]{C9667D}$68.8^{\scriptscriptstyle 72.0}_{\scriptscriptstyle 65.7}$ & \cellcolor[HTML]{66A487}$\mathbf{71.1}^{\scriptscriptstyle 74.0}_{\scriptscriptstyle 67.9}$ & \cellcolor[HTML]{D7EEAF}$70.3^{\scriptscriptstyle 73.5}_{\scriptscriptstyle 67.0}$ & \cellcolor[HTML]{D0EBAA}$70.3^{\scriptscriptstyle 73.5}_{\scriptscriptstyle 67.1}$ \\
& roc & \cellcolor[HTML]{66A487}$\mathbf{66.3}^{\scriptscriptstyle 69.5}_{\scriptscriptstyle 62.9}$ & \cellcolor[HTML]{ACDBA2}$66.1^{\scriptscriptstyle 69.7}_{\scriptscriptstyle 62.6}$ & \cellcolor[HTML]{71B992}$\underline{66.2}^{\scriptscriptstyle 69.5}_{\scriptscriptstyle 62.8}$ & \cellcolor[HTML]{C9667D}$65.5^{\scriptscriptstyle 68.6}_{\scriptscriptstyle 62.0}$ & \cellcolor[HTML]{91CF9D}$66.1^{\scriptscriptstyle 69.5}_{\scriptscriptstyle 62.8}$ \\
& noisy & \cellcolor[HTML]{8ACB9B}$63.6^{\scriptscriptstyle 67.2}_{\scriptscriptstyle 60.3}$ & \cellcolor[HTML]{66A487}$\mathbf{64.0}^{\scriptscriptstyle 67.9}_{\scriptscriptstyle 60.1}$ & \cellcolor[HTML]{88CA9B}$63.6^{\scriptscriptstyle 67.4}_{\scriptscriptstyle 60.2}$ & \cellcolor[HTML]{C9667D}$61.8^{\scriptscriptstyle 65.4}_{\scriptscriptstyle 58.3}$ & \cellcolor[HTML]{6BAD8C}$\underline{63.9}^{\scriptscriptstyle 67.5}_{\scriptscriptstyle 60.4}$ \\
\hline
\multirow{4}{*}{GleasonHD} & clean & \cellcolor[HTML]{F9AF92}$0.5^{\scriptscriptstyle 0.6}_{\scriptscriptstyle 0.5}$ & \cellcolor[HTML]{81C799}$\underline{0.6}^{\scriptscriptstyle 0.8}_{\scriptscriptstyle 0.5}$ & \cellcolor[HTML]{FFFCD4}$0.6^{\scriptscriptstyle 0.6}_{\scriptscriptstyle 0.5}$ & \cellcolor[HTML]{66A487}$\mathbf{0.7}^{\scriptscriptstyle 0.8}_{\scriptscriptstyle 0.6}$ & \cellcolor[HTML]{C9667D}$0.5^{\scriptscriptstyle 0.5}_{\scriptscriptstyle 0.4}$ \\
& roa & \cellcolor[HTML]{66A487}$\mathbf{0.5}^{\scriptscriptstyle 0.6}_{\scriptscriptstyle 0.5}$ & \cellcolor[HTML]{FEDAAA}$0.3^{\scriptscriptstyle 0.4}_{\scriptscriptstyle 0.3}$ & \cellcolor[HTML]{FFF0C0}$\underline{0.4}^{\scriptscriptstyle 0.4}_{\scriptscriptstyle 0.3}$ & \cellcolor[HTML]{F9AF92}$0.3^{\scriptscriptstyle 0.3}_{\scriptscriptstyle 0.3}$ & \cellcolor[HTML]{C9667D}$0.2^{\scriptscriptstyle 0.2}_{\scriptscriptstyle 0.2}$ \\
& roc & \cellcolor[HTML]{E7847E}$0.3^{\scriptscriptstyle 0.3}_{\scriptscriptstyle 0.2}$ & \cellcolor[HTML]{66A487}$\mathbf{0.5}^{\scriptscriptstyle 0.5}_{\scriptscriptstyle 0.4}$ & \cellcolor[HTML]{8CCC9C}$\underline{0.4}^{\scriptscriptstyle 0.5}_{\scriptscriptstyle 0.4}$ & \cellcolor[HTML]{CBE9A7}$0.4^{\scriptscriptstyle 0.4}_{\scriptscriptstyle 0.4}$ & \cellcolor[HTML]{C9667D}$0.2^{\scriptscriptstyle 0.3}_{\scriptscriptstyle 0.2}$ \\
& noisy & \cellcolor[HTML]{FFF2C2}$0.6^{\scriptscriptstyle 0.7}_{\scriptscriptstyle 0.5}$ & \cellcolor[HTML]{E9F6BB}$\underline{0.7}^{\scriptscriptstyle 0.9}_{\scriptscriptstyle 0.6}$ & \cellcolor[HTML]{F6A28C}$0.4^{\scriptscriptstyle 0.5}_{\scriptscriptstyle 0.4}$ & \cellcolor[HTML]{66A487}$\mathbf{1.0}^{\scriptscriptstyle 1.2}_{\scriptscriptstyle 0.9}$ & \cellcolor[HTML]{C9667D}$0.3^{\scriptscriptstyle 0.3}_{\scriptscriptstyle 0.2}$ \\
\hline
\multirow{4}{*}{MouseT} & clean & \cellcolor[HTML]{76C296}$\underline{90.4}^{\scriptscriptstyle 92.8}_{\scriptscriptstyle 88.1}$ & \cellcolor[HTML]{CAE8A6}$85.0^{\scriptscriptstyle 87.6}_{\scriptscriptstyle 82.3}$ & \cellcolor[HTML]{83C899}$89.6^{\scriptscriptstyle 92.1}_{\scriptscriptstyle 87.2}$ & \cellcolor[HTML]{C9667D}$66.0^{\scriptscriptstyle 69.1}_{\scriptscriptstyle 62.8}$ & \cellcolor[HTML]{66A487}$\mathbf{93.2}^{\scriptscriptstyle 95.3}_{\scriptscriptstyle 91.1}$ \\
& roa & \cellcolor[HTML]{66A487}$\mathbf{95.9}^{\scriptscriptstyle 97.5}_{\scriptscriptstyle 94.2}$ & \cellcolor[HTML]{F29A88}$87.1^{\scriptscriptstyle 89.5}_{\scriptscriptstyle 84.5}$ & \cellcolor[HTML]{FFF1C1}$89.8^{\scriptscriptstyle 92.3}_{\scriptscriptstyle 87.3}$ & \cellcolor[HTML]{C9667D}$85.3^{\scriptscriptstyle 87.6}_{\scriptscriptstyle 83.2}$ & \cellcolor[HTML]{F7FCCE}$\underline{90.9}^{\scriptscriptstyle 93.1}_{\scriptscriptstyle 88.9}$ \\
& roc & \cellcolor[HTML]{9AD39F}$88.1^{\scriptscriptstyle 90.8}_{\scriptscriptstyle 85.6}$ & \cellcolor[HTML]{C9667D}$60.9^{\scriptscriptstyle 64.0}_{\scriptscriptstyle 57.9}$ & \cellcolor[HTML]{73BC93}$\underline{91.4}^{\scriptscriptstyle 93.4}_{\scriptscriptstyle 89.3}$ & \cellcolor[HTML]{A0D6A1}$87.7^{\scriptscriptstyle 90.2}_{\scriptscriptstyle 85.3}$ & \cellcolor[HTML]{66A487}$\mathbf{94.2}^{\scriptscriptstyle 96.3}_{\scriptscriptstyle 91.9}$ \\
& noisy & \cellcolor[HTML]{D6EEAE}$89.0^{\scriptscriptstyle 91.4}_{\scriptscriptstyle 86.7}$ & \cellcolor[HTML]{C9667D}$79.5^{\scriptscriptstyle 82.3}_{\scriptscriptstyle 76.4}$ & \cellcolor[HTML]{91CF9D}$\underline{91.6}^{\scriptscriptstyle 93.9}_{\scriptscriptstyle 89.0}$ & \cellcolor[HTML]{FEE7B4}$85.0^{\scriptscriptstyle 87.3}_{\scriptscriptstyle 82.4}$ & \cellcolor[HTML]{66A487}$\mathbf{93.9}^{\scriptscriptstyle 95.7}_{\scriptscriptstyle 91.9}$ \\
\hline
\multirow{4}{*}{MMIS} & clean & \cellcolor[HTML]{FBBA97}$81.2^{\scriptscriptstyle 86.3}_{\scriptscriptstyle 76.2}$ & \cellcolor[HTML]{C9667D}$78.5^{\scriptscriptstyle 83.6}_{\scriptscriptstyle 73.4}$ & \cellcolor[HTML]{66A487}$\mathbf{89.3}^{\scriptscriptstyle 92.5}_{\scriptscriptstyle 86.1}$ & \cellcolor[HTML]{D16E7D}$78.8^{\scriptscriptstyle 84.8}_{\scriptscriptstyle 73.2}$ & \cellcolor[HTML]{68A889}$\underline{89.1}^{\scriptscriptstyle 93.8}_{\scriptscriptstyle 83.9}$ \\
& roa & \cellcolor[HTML]{F7A58D}$83.2^{\scriptscriptstyle 87.2}_{\scriptscriptstyle 79.1}$ & \cellcolor[HTML]{FEEFBE}$86.3^{\scriptscriptstyle 92.5}_{\scriptscriptstyle 80.3}$ & \cellcolor[HTML]{66A487}$\mathbf{94.2}^{\scriptscriptstyle 97.2}_{\scriptscriptstyle 91.1}$ & \cellcolor[HTML]{C9667D}$80.6^{\scriptscriptstyle 85.3}_{\scriptscriptstyle 76.4}$ & \cellcolor[HTML]{D7EEAF}$\underline{89.5}^{\scriptscriptstyle 92.7}_{\scriptscriptstyle 85.5}$ \\
& roc & \cellcolor[HTML]{E7F5B9}$83.4^{\scriptscriptstyle 89.4}_{\scriptscriptstyle 77.7}$ & \cellcolor[HTML]{C9667D}$76.4^{\scriptscriptstyle 82.1}_{\scriptscriptstyle 70.8}$ & \cellcolor[HTML]{66A487}$\mathbf{87.9}^{\scriptscriptstyle 92.6}_{\scriptscriptstyle 83.4}$ & \cellcolor[HTML]{F4FAC9}$82.7^{\scriptscriptstyle 89.1}_{\scriptscriptstyle 76.7}$ & \cellcolor[HTML]{7FC698}$\underline{86.5}^{\scriptscriptstyle 90.2}_{\scriptscriptstyle 82.7}$ \\
& noisy & \cellcolor[HTML]{E27E7D}$73.6^{\scriptscriptstyle 76.7}_{\scriptscriptstyle 70.1}$ & \cellcolor[HTML]{C9667D}$72.4^{\scriptscriptstyle 76.9}_{\scriptscriptstyle 68.3}$ & \cellcolor[HTML]{66A487}$\mathbf{86.3}^{\scriptscriptstyle 90.1}_{\scriptscriptstyle 82.6}$ & \cellcolor[HTML]{F49D8A}$74.8^{\scriptscriptstyle 80.3}_{\scriptscriptstyle 69.4}$ & \cellcolor[HTML]{9ED5A0}$\underline{83.6}^{\scriptscriptstyle 88.6}_{\scriptscriptstyle 78.6}$ \\
\hline
\multirow{4}{*}{MMIA} & clean & \cellcolor[HTML]{9CD4A0}$32.7^{\scriptscriptstyle 35.4}_{\scriptscriptstyle 30.2}$ & \cellcolor[HTML]{C9667D}$24.1^{\scriptscriptstyle 26.3}_{\scriptscriptstyle 21.9}$ & \cellcolor[HTML]{7CC498}$\underline{33.5}^{\scriptscriptstyle 36.3}_{\scriptscriptstyle 30.6}$ & \cellcolor[HTML]{F49F8A}$26.0^{\scriptscriptstyle 28.2}_{\scriptscriptstyle 23.8}$ & \cellcolor[HTML]{66A487}$\mathbf{34.7}^{\scriptscriptstyle 37.3}_{\scriptscriptstyle 32.2}$ \\
& roa & \cellcolor[HTML]{FDCC9F}$25.6^{\scriptscriptstyle 27.8}_{\scriptscriptstyle 23.5}$ & \cellcolor[HTML]{C9667D}$22.9^{\scriptscriptstyle 24.9}_{\scriptscriptstyle 21.0}$ & \cellcolor[HTML]{CEEAA8}$\underline{29.3}^{\scriptscriptstyle 31.3}_{\scriptscriptstyle 27.4}$ & \cellcolor[HTML]{ECF7BE}$28.3^{\scriptscriptstyle 30.7}_{\scriptscriptstyle 25.8}$ & \cellcolor[HTML]{66A487}$\mathbf{32.2}^{\scriptscriptstyle 35.4}_{\scriptscriptstyle 29.3}$ \\
& roc & \cellcolor[HTML]{C9667D}$18.9^{\scriptscriptstyle 20.5}_{\scriptscriptstyle 17.3}$ & \cellcolor[HTML]{FBBB97}$20.1^{\scriptscriptstyle 22.6}_{\scriptscriptstyle 17.9}$ & \cellcolor[HTML]{E5F4B7}$\underline{21.7}^{\scriptscriptstyle 23.3}_{\scriptscriptstyle 20.1}$ & \cellcolor[HTML]{E6F4B8}$21.7^{\scriptscriptstyle 24.1}_{\scriptscriptstyle 19.7}$ & \cellcolor[HTML]{66A487}$\mathbf{23.4}^{\scriptscriptstyle 26.0}_{\scriptscriptstyle 21.2}$ \\
& noisy & \cellcolor[HTML]{F49D8A}$16.7^{\scriptscriptstyle 18.3}_{\scriptscriptstyle 15.2}$ & \cellcolor[HTML]{DD797D}$16.1^{\scriptscriptstyle 17.7}_{\scriptscriptstyle 14.6}$ & \cellcolor[HTML]{8ACB9B}$\underline{20.5}^{\scriptscriptstyle 22.2}_{\scriptscriptstyle 18.8}$ & \cellcolor[HTML]{C9667D}$15.8^{\scriptscriptstyle 17.1}_{\scriptscriptstyle 14.6}$ & \cellcolor[HTML]{66A487}$\mathbf{21.3}^{\scriptscriptstyle 23.7}_{\scriptscriptstyle 19.1}$ \\
\hline
\hline
\end{tabular}

\label{tab:bootstrap_fgbginstancef1_results}
\end{table*}

\begin{figure}[h!]
\centering
\includegraphics[width=\textwidth]{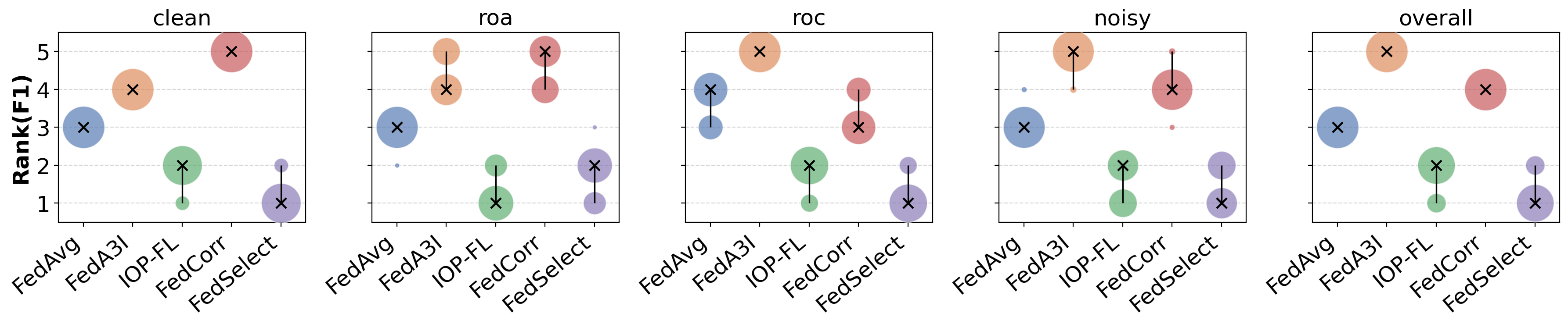}
\caption{Rank stability of foreground-background instance-level F1 segmentation performances based on bootstrapping across client-noise scenarios and methods on LIDC, GleasonHD, MouseT, MMIS and MMIA. Lower ranks indicate better performances, bubble size reflects the ranking frequency across datasets.}
\label{fig:app_ranking_fgbginstancef1}
\end{figure}

\clearpage

\subsubsection{Confusion-based segmentation label noise}
\label{app2:noisespecific_clsconf}

\begin{figure}[h!]
\centering
\includegraphics[width=\textwidth]{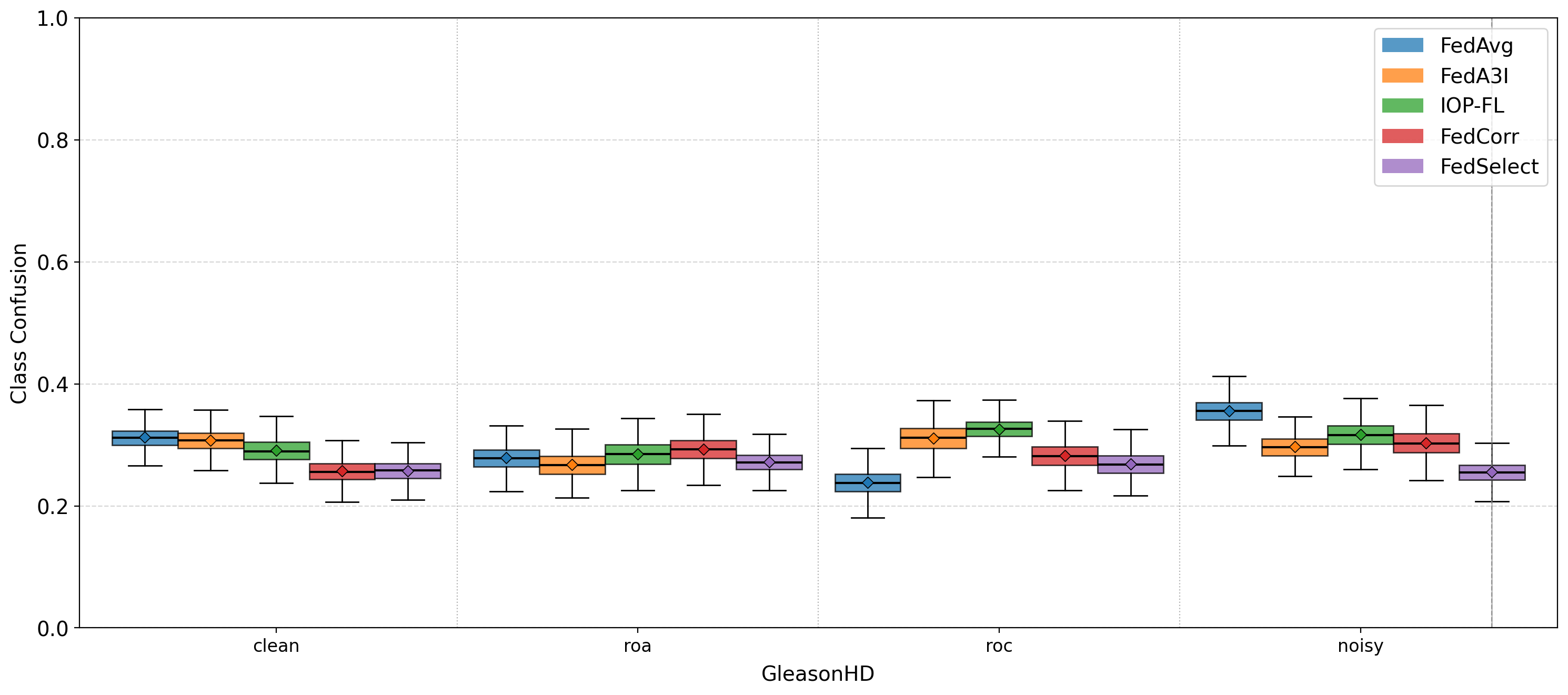}
\caption{Bootstrap Class confusion score distributions of all benchmarked methods across client-noise scenarios and all datasets featuring confusion-based noise (GleasonHD). Black bars denote medians, and diamonds denote means.}
\label{fig:app_classconfusion}
\end{figure}

\begin{figure}[h!]
\centering
\includegraphics[width=\textwidth]{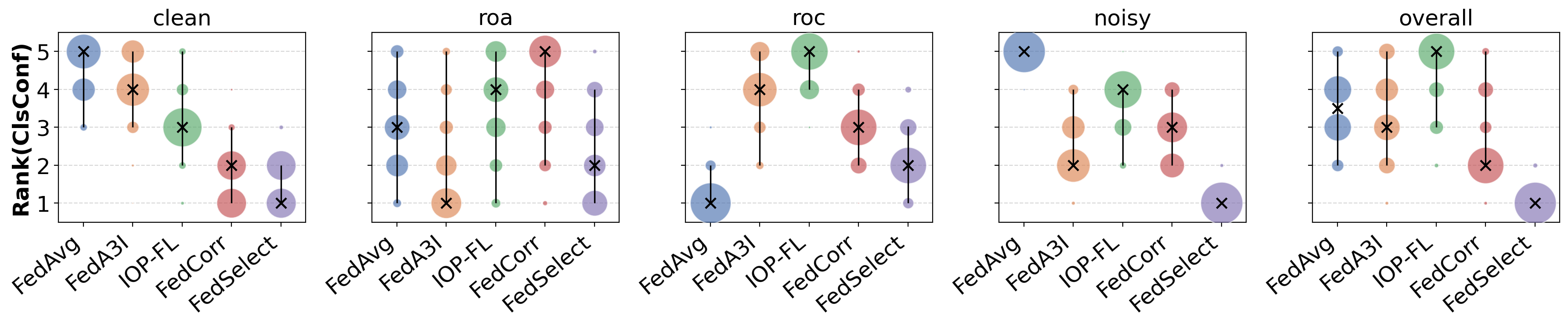}
\caption{Rank stability of Class confusion segmentation performances based on bootstrapping across client-noise scenarios and methods on GleasonHD. Lower ranks indicate better performances, bubble size reflects the ranking frequency across datasets.}
\label{fig:app_ranking_classconfusion}
\end{figure}

Confusion-related segmentation label noise was assessed using the voxel-level class confusion metric on datasets featuring this noise type. Detailed results are reported in Table~\ref{tab:bootstrap_classconfusion_results} and Figures~\ref{fig:app_classconfusion},~\ref{fig:app_ranking_classconfusion}.

\begin{table*}[h!]
\centering

\caption{Mean validation ClsConf values with 95\% percentile bootstrap confidence intervals for each dataset, noise scenario, and method.}

\small
\begin{tabular}{llccccc}
\hline
\hline
\textit{Dataset} & \textit{Scenario} & \textbf{FedAvg} & \textbf{FedA3I} & \textbf{IOP-FL} & \textbf{FedCorr} & \textbf{FedSelect} \\
\hline
\multirow{4}{*}{GleasonHD} & clean & \cellcolor[HTML]{C9667D}$31.2^{\scriptscriptstyle 35.2}_{\scriptscriptstyle 27.9}$ & \cellcolor[HTML]{E37F7D}$30.8^{\scriptscriptstyle 34.6}_{\scriptscriptstyle 27.3}$ & \cellcolor[HTML]{FEE9B6}$29.1^{\scriptscriptstyle 33.5}_{\scriptscriptstyle 25.0}$ & \cellcolor[HTML]{66A487}$\mathbf{25.7}^{\scriptscriptstyle 30.0}_{\scriptscriptstyle 22.0}$ & \cellcolor[HTML]{67A788}$\underline{25.8}^{\scriptscriptstyle 29.1}_{\scriptscriptstyle 22.4}$ \\
& roa & \cellcolor[HTML]{F3FAC8}$27.9^{\scriptscriptstyle 32.4}_{\scriptscriptstyle 23.9}$ & \cellcolor[HTML]{66A487}$\mathbf{26.8}^{\scriptscriptstyle 31.5}_{\scriptscriptstyle 22.7}$ & \cellcolor[HTML]{FECFA1}$28.5^{\scriptscriptstyle 32.9}_{\scriptscriptstyle 24.3}$ & \cellcolor[HTML]{C9667D}$29.3^{\scriptscriptstyle 33.7}_{\scriptscriptstyle 25.2}$ & \cellcolor[HTML]{93D09D}$\underline{27.2}^{\scriptscriptstyle 31.1}_{\scriptscriptstyle 23.7}$ \\
& roc & \cellcolor[HTML]{66A487}$\mathbf{23.8}^{\scriptscriptstyle 28.0}_{\scriptscriptstyle 20.0}$ & \cellcolor[HTML]{F29A88}$31.1^{\scriptscriptstyle 35.9}_{\scriptscriptstyle 26.5}$ & \cellcolor[HTML]{C9667D}$32.6^{\scriptscriptstyle 36.3}_{\scriptscriptstyle 29.0}$ & \cellcolor[HTML]{FFFED7}$28.3^{\scriptscriptstyle 32.6}_{\scriptscriptstyle 24.3}$ & \cellcolor[HTML]{D7EEAF}$\underline{26.9}^{\scriptscriptstyle 30.9}_{\scriptscriptstyle 23.1}$ \\
& noisy & \cellcolor[HTML]{C9667D}$35.6^{\scriptscriptstyle 39.7}_{\scriptscriptstyle 31.7}$ & \cellcolor[HTML]{EBF7BD}$\underline{29.7}^{\scriptscriptstyle 33.8}_{\scriptscriptstyle 26.0}$ & \cellcolor[HTML]{FEE9B6}$31.7^{\scriptscriptstyle 36.1}_{\scriptscriptstyle 27.6}$ & \cellcolor[HTML]{F9FDD1}$30.3^{\scriptscriptstyle 34.7}_{\scriptscriptstyle 26.1}$ & \cellcolor[HTML]{66A487}$\mathbf{25.6}^{\scriptscriptstyle 29.5}_{\scriptscriptstyle 22.3}$ \\
\hline
\hline
\end{tabular}

\label{tab:bootstrap_classconfusion_results}
\end{table*}

\clearpage

\subsection{Statistical evaluation of segmentation performances}
\label{app3:statistics}

All tests compare each FNLL method against FedAvg using one-sided paired Wilcoxon signed-rank tests on case-level scores (Table~\ref{tab:fnll_wilcoxon_per_metric_per_scenario}). For each metric, scenario, and method, scores are matched to FedAvg by case identifier within each dataset and pooled across datasets. Positive $\Delta$ values indicate improvement over FedAvg, corresponding to lower values for HD95 and class confusion. Dice uses all datasets, whereas noise-sensitive metrics use only datasets featuring the respective noise type: all datasets for HD95, LIDC, GleasonHD, MouseT, MMIS, and MMIA for foreground-background instance-level F1, and GleasonHD for class confusion. Holm--Bonferroni correction is applied across the four method comparisons per test family; underlined $p$-values indicate uncorrected significance, and bold corrected $p$-values indicate significance after correction.

\begin{longtable}{lllrrrrr}
\caption{Per-metric, per-scenario Wilcoxon signed-rank tests against FedAvg.}\label{tab:fnll_wilcoxon_per_metric_per_scenario}\\
\toprule
Metric & Scenario & Method & $n$ & $\Delta_{mean}$ & $\Delta_{median}$ & $p$ & $p_{Holm}$ \\
\midrule
\endfirsthead
\caption[]{Per-metric, per-scenario Wilcoxon signed-rank tests against FedAvg. (continued)}\\
\toprule
Metric & Scenario & Method & $n$ & $\Delta_{mean}$ & $\Delta_{median}$ & $p$ & $p_{Holm}$ \\
\midrule
\endhead
\midrule
\multicolumn{8}{r}{Continued on next page} \\
\endfoot
\bottomrule
\endlastfoot
\multirow{16}{*}{Dice} & \multirow{4}{*}{clean} & FedA3I & 3261 & -0.0330 & -0.0081 & 1.0000 & 1.0000 \\
 &  & IOP-FL & 3263 & -0.0048 & +0.0000 & 0.6592 & 1.0000 \\
 &  & FedCorr & 3266 & -0.0379 & -0.0085 & 1.0000 & 1.0000 \\
 &  & FedSelect & 3266 & -0.0022 & +0.0000 & 0.9991 & 1.0000 \\
\cmidrule(lr){2-8}
 & \multirow{4}{*}{roa} & FedA3I & 3266 & -0.0129 & -0.0028 & 1.0000 & 1.0000 \\
 &  & IOP-FL & 3265 & +0.0097 & +0.0000 & \underline{1.14e-06} & \textbf{4.55e-06} \\
 &  & FedCorr & 3267 & -0.0101 & -0.0035 & 1.0000 & 1.0000 \\
 &  & FedSelect & 3266 & +0.0029 & +0.0000 & 0.8275 & 1.0000 \\
\cmidrule(lr){2-8}
 & \multirow{4}{*}{roc} & FedA3I & 3257 & -0.0149 & -0.0021 & 1.0000 & 1.0000 \\
 &  & IOP-FL & 3262 & +0.0003 & +0.0000 & \underline{0.0106} & \textbf{0.0423} \\
 &  & FedCorr & 3264 & -0.0123 & -0.0011 & 1.0000 & 1.0000 \\
 &  & FedSelect & 3258 & +0.0038 & +0.0000 & 0.7754 & 1.0000 \\
\cmidrule(lr){2-8}
 & \multirow{4}{*}{noisy} & FedA3I & 3250 & -0.0157 & -0.0021 & 1.0000 & 1.0000 \\
 &  & IOP-FL & 3262 & +0.0078 & +0.0000 & \underline{1.07e-10} & \textbf{4.26e-10} \\
 &  & FedCorr & 3259 & -0.0193 & -0.0016 & 1.0000 & 1.0000 \\
 &  & FedSelect & 3260 & +0.0032 & +0.0000 & 0.3777 & 1.0000 \\
\midrule
\multirow{16}{*}{HD95} & \multirow{4}{*}{clean} & FedA3I & 3268 & -17.9889 & +0.0000 & 1.0000 & 1.0000 \\
 &  & IOP-FL & 3268 & -6.0093 & +0.0000 & 0.9931 & 1.0000 \\
 &  & FedCorr & 3268 & -15.5687 & +0.0000 & 1.0000 & 1.0000 \\
 &  & FedSelect & 3268 & +3.4931 & +0.0000 & 0.5947 & 1.0000 \\
\cmidrule(lr){2-8}
 & \multirow{4}{*}{roa} & FedA3I & 3268 & -4.9229 & +0.0000 & 1.0000 & 1.0000 \\
 &  & IOP-FL & 3268 & +1.4570 & +0.0000 & 0.1147 & 0.4586 \\
 &  & FedCorr & 3268 & -3.1700 & +0.0000 & 1.0000 & 1.0000 \\
 &  & FedSelect & 3268 & -0.4616 & +0.0000 & 0.8167 & 1.0000 \\
\cmidrule(lr){2-8}
 & \multirow{4}{*}{roc} & FedA3I & 3268 & -11.7246 & +0.0000 & 1.0000 & 1.0000 \\
 &  & IOP-FL & 3268 & -4.9786 & +0.0000 & \underline{0.0084} & \textbf{0.0253} \\
 &  & FedCorr & 3268 & -5.7519 & +0.0000 & 0.9856 & 1.0000 \\
 &  & FedSelect & 3268 & +3.9923 & +0.0000 & \underline{0.0023} & \textbf{0.0091} \\
\cmidrule(lr){2-8}
 & \multirow{4}{*}{noisy} & FedA3I & 3268 & -10.0181 & +0.0000 & 1.0000 & 1.0000 \\
 &  & IOP-FL & 3268 & -9.0351 & +0.0000 & \underline{0.0438} & 0.1751 \\
 &  & FedCorr & 3268 & -11.7081 & +0.0000 & 1.0000 & 1.0000 \\
 &  & FedSelect & 3268 & -0.7138 & +0.0000 & 0.1112 & 0.3337 \\
\midrule
\multirow{16}{*}{F1} & \multirow{4}{*}{clean} & FedA3I & 3261 & -0.0461 & +0.0000 & 1.0000 & 1.0000 \\
 &  & IOP-FL & 3263 & -0.0018 & +0.0000 & 0.5221 & 1.0000 \\
 &  & FedCorr & 3266 & -0.0610 & +0.0000 & 1.0000 & 1.0000 \\
 &  & FedSelect & 3266 & +0.0077 & +0.0000 & \underline{0.0015} & \textbf{0.0058} \\
\cmidrule(lr){2-8}
 & \multirow{4}{*}{roa} & FedA3I & 3266 & -0.0344 & +0.0000 & 1.0000 & 1.0000 \\
 &  & IOP-FL & 3265 & +0.0021 & +0.0000 & 0.1803 & 0.5408 \\
 &  & FedCorr & 3267 & -0.0287 & +0.0000 & 1.0000 & 1.0000 \\
 &  & FedSelect & 3266 & +0.0072 & +0.0000 & \underline{2.18e-06} & \textbf{8.74e-06} \\
\cmidrule(lr){2-8}
 & \multirow{4}{*}{roc} & FedA3I & 3257 & -0.0401 & +0.0000 & 1.0000 & 1.0000 \\
 &  & IOP-FL & 3262 & -0.0005 & +0.0000 & 0.3762 & 1.0000 \\
 &  & FedCorr & 3264 & -0.0201 & +0.0000 & 0.9162 & 1.0000 \\
 &  & FedSelect & 3258 & +0.0132 & +0.0000 & \underline{2.43e-09} & \textbf{9.71e-09} \\
\cmidrule(lr){2-8}
 & \multirow{4}{*}{noisy} & FedA3I & 3250 & -0.0346 & +0.0000 & 1.0000 & 1.0000 \\
 &  & IOP-FL & 3262 & +0.0075 & +0.0000 & \underline{1.57e-08} & \textbf{6.27e-08} \\
 &  & FedCorr & 3259 & -0.0314 & +0.0000 & 1.0000 & 1.0000 \\
 &  & FedSelect & 3260 & +0.0110 & +0.0000 & \underline{1.31e-07} & \textbf{3.93e-07} \\
\midrule
\multirow{16}{*}{ClsConf} & \multirow{4}{*}{clean} & FedA3I & 738 & +0.0030 & -0.0055 & 1.0000 & 1.0000 \\
 &  & IOP-FL & 738 & +0.0075 & +0.0000 & 0.2027 & 0.8107 \\
 &  & FedCorr & 738 & +0.0068 & -0.0039 & 1.0000 & 1.0000 \\
 &  & FedSelect & 738 & +0.0166 & -0.0005 & 1.0000 & 1.0000 \\
\cmidrule(lr){2-8}
 & \multirow{4}{*}{roa} & FedA3I & 738 & -0.0050 & -0.0018 & 1.0000 & 1.0000 \\
 &  & IOP-FL & 738 & -0.0003 & +0.0000 & 0.8516 & 1.0000 \\
 &  & FedCorr & 738 & -0.0024 & -0.0002 & 1.0000 & 1.0000 \\
 &  & FedSelect & 738 & +0.0006 & -0.0021 & 1.0000 & 1.0000 \\
\cmidrule(lr){2-8}
 & \multirow{4}{*}{roc} & FedA3I & 738 & -0.0214 & +0.0000 & 0.9885 & 1.0000 \\
 &  & IOP-FL & 738 & -0.0281 & -0.0006 & 0.9995 & 1.0000 \\
 &  & FedCorr & 738 & -0.0109 & +0.0000 & 0.9326 & 1.0000 \\
 &  & FedSelect & 738 & -0.0133 & -0.0004 & 1.0000 & 1.0000 \\
\cmidrule(lr){2-8}
 & \multirow{4}{*}{noisy} & FedA3I & 738 & +0.0101 & +0.0000 & 0.1729 & 0.5187 \\
 &  & IOP-FL & 738 & +0.0038 & +0.0000 & \underline{0.0076} & \textbf{0.0305} \\
 &  & FedCorr & 738 & +0.0011 & -0.0000 & 0.9888 & 0.9888 \\
 &  & FedSelect & 738 & +0.0194 & +0.0000 & 0.2477 & 0.5187 \\
\end{longtable}

\clearpage

To increase statistical power while preserving interpretability, we additionally pool case-level paired observations across datasets and client-noise scenarios within each metric (Table~\ref{tab:fnll_wilcoxon_pooled_dataset_x_scenario_per_metric}). The test direction, effect definition, and correction procedure remain identical, with Holm-Bonferroni correction applied across the four method comparisons per metric.

\begin{table}[h!]
\centering
\caption{Pooled dataset-by-scenario Wilcoxon signed-rank tests against FedAvg.}
\label{tab:fnll_wilcoxon_pooled_dataset_x_scenario_per_metric}
\begin{tabular}{lllrrrrr}
\toprule
Metric & Scenario & Method & $n$ & $\Delta_{mean}$ & $\Delta_{median}$ & $p$ & $p_{Holm}$ \\
\midrule
\multirow{4}{*}{Dice} & \multirow{4}{*}{ALL} & FedA3I & 13034 & -0.0191 & -0.0038 & 1.0000 & 1.0000 \\
 &  & IOP-FL & 13052 & +0.0032 & +0.0000 & \underline{3.44e-11} & \textbf{1.38e-10} \\
 &  & FedCorr & 13056 & -0.0199 & -0.0036 & 1.0000 & 1.0000 \\
 &  & FedSelect & 13050 & +0.0019 & +0.0000 & 0.9874 & 1.0000 \\
\midrule
\multirow{4}{*}{HD95} & \multirow{4}{*}{ALL} & FedA3I & 13072 & -11.1636 & +0.0000 & 1.0000 & 1.0000 \\
 &  & IOP-FL & 13072 & -4.6415 & +0.0000 & 0.0772 & 0.2708 \\
 &  & FedCorr & 13072 & -9.0497 & +0.0000 & 1.0000 & 1.0000 \\
 &  & FedSelect & 13072 & +1.5775 & +0.0000 & 0.0677 & 0.2708 \\
\midrule
\multirow{4}{*}{F1} & \multirow{4}{*}{ALL} & FedA3I & 13034 & -0.0388 & +0.0000 & 1.0000 & 1.0000 \\
 &  & IOP-FL & 13052 & +0.0018 & +0.0000 & \underline{5.17e-04} & \textbf{0.0016} \\
 &  & FedCorr & 13056 & -0.0353 & +0.0000 & 1.0000 & 1.0000 \\
 &  & FedSelect & 13050 & +0.0098 & +0.0000 & \underline{5.39e-21} & \textbf{2.16e-20} \\
\midrule
\multirow{4}{*}{ClsConf} & \multirow{4}{*}{ALL} & FedA3I & 2952 & -0.0033 & -0.0003 & 1.0000 & 1.0000 \\
 &  & IOP-FL & 2952 & -0.0043 & +0.0000 & 0.7221 & 1.0000 \\
 &  & FedCorr & 2952 & -0.0013 & -0.0000 & 1.0000 & 1.0000 \\
 &  & FedSelect & 2952 & +0.0058 & -0.0003 & 1.0000 & 1.0000 \\
\bottomrule
\end{tabular}
\end{table}

Given the large number of paired case-level observations, the Wilcoxon test primarily reflects the consistency of directional improvements rather than their magnitude. Consequently, statistically significant results indicate systematic per-case gains over FedAvg, whereas non-significant results often reflect heterogeneous or dataset-dependent behavior.

\clearpage

\end{appendices}


\bibliography{sn-bibliography}

\end{document}